%% file: 2024-wsdm-fp-reign.tex
\documentclass[sigconf,natbib=true,anonymous=false]{acmart}

\usepackage{balance}
\usepackage{hyperref}
\usepackage{latexsym}
\usepackage{graphicx}
\graphicspath{{./images/}}
\usepackage{booktabs} 
\usepackage{color}  
\usepackage{amsmath}  
\usepackage{subcaption}
\usepackage{caption}
\usepackage{tikz}
\usepackage{colortbl} 
\usepackage{framed}
\usepackage{multirow}
\usepackage{multicol}
\usepackage{url}
\usepackage{balance}
\usepackage{verbatim}
\usepackage{cancel}
\usepackage{xspace} 
\usepackage[ruled,linesnumbered,vlined]{algorithm2e}
\usepackage{bbold} 
\usepackage{arydshln}

\newcommand{\struct}[1]{\texttt{\small #1}}
\newcommand{\utterance}[1]{\textit{#1}}
\newcommand{\phrase}[1]{\textit{``#1''}}

\newcommand{\myparagraph}[1]{\noindent \textbf{#1}.}

\newenvironment{Snugshade}[1][236,236,236]{
	\setlength{\itemsep}{0pt}
	\setlength{\parsep}{0pt}
	\setlength{\topsep}{0pt}
	\setlength{\partopsep}{0pt}
	\setlength{\leftmargin}{1.5em}
	\setlength{\labelwidth}{0em}
	\setlength{\labelsep}{0em} 
	\setlength{\parskip}{0pt}
	\definecolor{shadecolor}{RGB}{#1}%
	\begin{snugshade}
	}{%
	\end{snugshade}%
}

\newcommand{\reign}{\textsc{Reign}\xspace}
\newcommand{\conquer}{\textsc{Conquer}\xspace}

\newcommand{\convmix}{\textsc{ConvMix}\xspace}

\newcommand{\explaignn}{\textsc{Explaignn}\xspace}
\newcommand{\convquestions}{\textsc{ConvQuestions}\xspace}

\newcommand{\squishlist}{
	\begin{list}{$\bullet$}
		{ \setlength{\itemsep}{0pt}
			\setlength{\parsep}{3pt}
			\setlength{\topsep}{3pt}
			\setlength{\partopsep}{0pt}
			\setlength{\leftmargin}{1.5em}
			\setlength{\labelwidth}{1em}
			\setlength{\labelsep}{0.5em} } }
	\newcommand{\squishend}{
\end{list}  }

\AtBeginDocument{%
  \providecommand\BibTeX{{%
    \normalfont B\kern-0.5em{\scshape i\kern-0.25em b}\kern-0.8em\TeX}}}

\setcopyright{acmcopyright}
\copyrightyear{2024}
\acmYear{2024}
\acmDOI{XXXXXXX.XXXXXXX}

\acmConference[WSDM '24]{Proceedings of the 17th ACM International Conference on Web Search and Data Mining}{4 -- 8 March 2024}{M\'{e}rida, M\'{e}xico}

\begin{document}

\title{Robust Training for Conversational Question Answering Models with Reinforced Reformulation Generation}

	\author{Magdalena Kaiser}
	\affiliation{
		\institution{Max Planck Institute for Informatics
		\\ Saarland Informatics Campus}
        \country{Germany}}
	\email{mkaiser@mpi-inf.mpg.de}
	
	\author{Rishiraj Saha Roy}
	\affiliation{
		\institution{Max Planck Institute for Informatics
		\\ Saarland Informatics Campus}
        \country{Germany}}
	\email{rishiraj@mpi-inf.mpg.de}
	
	\author{Gerhard Weikum}
	\affiliation{
		\institution{Max Planck Institute for Informatics
		\\ Saarland Informatics Campus}
        \country{Germany}}
	\email{weikum@mpi-inf.mpg.de}

\renewcommand{\shortauthors}{Kaiser et al.}


\input{sections/00-abstract}

\settopmatter{printacmref=true, printccs=true, printfolios=true}

\ccsdesc[500]{Information systems~Question answering}

\keywords{Question answering, Knowledge graphs, Conversations, Reformulations, Reinforcement learning}

\maketitle

\input{sections/01-introduction}
\input{sections/02-concepts.tex}
\input{sections/03-method}
\input{sections/04-experiments}
\input{sections/05-results}

\input{sections/06-related}
\input{sections/07-conclusion}


\input{sections/08-ethics}

\balance
\bibliographystyle{ACM-Reference-Format}
\bibliography{2024-wsdm-fp-reign}

\end{document}

%% file: sections/00-abstract.tex
\begin{abstract}
Models for conversational question answering (ConvQA) over knowledge graphs (KGs) are usually trained and tested on benchmarks of gold QA pairs. This implies that training is limited to surface forms seen in the respective datasets, and evaluation is on a small set of held-out questions. Through our proposed framework \reign, we take several steps to remedy this restricted learning setup. First, we systematically generate reformulations of training questions to increase robustness of models to surface form variations. This is a particularly challenging problem, given the incomplete nature of such questions. Second, we guide ConvQA models towards higher performance by feeding it only those reformulations that help improve their answering quality, using deep reinforcement learning. Third, we demonstrate the viability of training major model components on one benchmark and applying them zero-shot to another. Finally, for a rigorous evaluation of robustness for trained models, we use and release large numbers of diverse reformulations generated by prompting GPT for benchmark test sets (resulting in 20x increase in sizes). Our findings show that ConvQA models with robust training via reformulations significantly outperform those with standard training from gold QA pairs only.
\end{abstract}

%% file: sections/01-introduction.tex
\section{Introduction}
\label{sec:intro}

\myparagraph{Motivation} Answering questions about entities, 
powered by curated knowledge graphs (KGs) at the backend, is a vital component of Web search~\cite{bast2015more,linjordet2022would,saharoy2022question,yih2015semantic}. 
Nowadays, users' information needs are increasingly being expressed as a conversation, in a sequence of questions and answers $\langle Q_t, A_t \rangle$, over turns $\{t\}$~\cite{zamani2022conversational,dalton2022conversational,owoicho2023exploiting}:
\begin{Snugshade}
	$Q_1$: \utterance{What's the 2022 LOTR TV series called?}\\
    \indent $A_1$: \struct{The Rings of Power (TROP)}		

	$Q_2$: \utterance{TROP airing on?}\\
	\indent $A_2$: \struct{Amazon Prime Video}

 	$Q_3$: \utterance{Which actor plays Isildur in the series?}\\
	\indent $A_3$: \struct{Maxim Baldry}

    $Q_4$: \utterance{And who in the Jackson trilogy?}\\
    \indent $A_4$: \struct{Harry Sinclair} 
    
    $Q_5$: \utterance{When did the series start?} ...
\end{Snugshade}
A conversation over a KG contains a set of entities (\phrase{The Lord of the Rings: The Rings of Power}, \phrase{Amazon Prime Video}), their relationships (\phrase{aired on}), and types (\phrase{TV series, video streaming service}). 
In ConvQA, users omit parts of the context in several follow-up turns ($Q_3 - Q_5$), and use ad hoc style ($Q_2$)~\cite{kaiser2020conversational,dalton2019cast,radlinski2017theoretical,dalton2022conversational,guo2018dialog,shen2019multi}.
A part of the intent being left implicit, coupled with the use of informal language, make the answering of conversational questions more challenging than complete ones tackled in older and more established branches of QA~\cite{voorhees1999trec,hirschman2001natural,saharoy2022question,usbeck2017open}. 
ConvQA 
has high contemporary interest~\cite{kacupaj2022contrastive,christmann2023explainable,qu2020open,li2022mmcoqa,vakulenko2021question}, spurred on to a big extent by systems like ChatGPT that support a conversational interface.

\myparagraph{Limitations of state-of-the-art} We quantify \textit{robustness} in QA in terms of the number of distinct question formulations of a given intent, that a QA model can answer correctly: the higher this number, the more robust the model.
Methods for conversational question answering (ConvQA) over KGs are usually trained and evaluated on benchmarks of gold-standard $\langle$question, answer$\rangle$ pairs~\cite{saha2018complex,shen2019multi,guo2018dialog,christmann2022conversational}.
Such a paradigm limits \textit{robust learning} by being restricted to question formulations roughly seen during training time.
One approach in QA to demonstrate generalizability is to train and evaluate models on multiple benchmarks~\cite{kacupaj2022contrastive,marion2021structured,lan2021modeling}.
This only addresses the problem partially: the training and evaluation are still limited to surface forms seen in any of the benchmarks. 
A particular aspect of existing benchmarks, that is attributable to their construction choices via graph sampling~\cite{saha2018complex} or crowdsourcing guidelines~\cite{christmann2019look,christmann2022conversational}, is that they often do not contain sloppy question formulations that could be asked by real users in the wild.

\begin{figure*} [t]
	\centering
    \includegraphics[width=\textwidth]{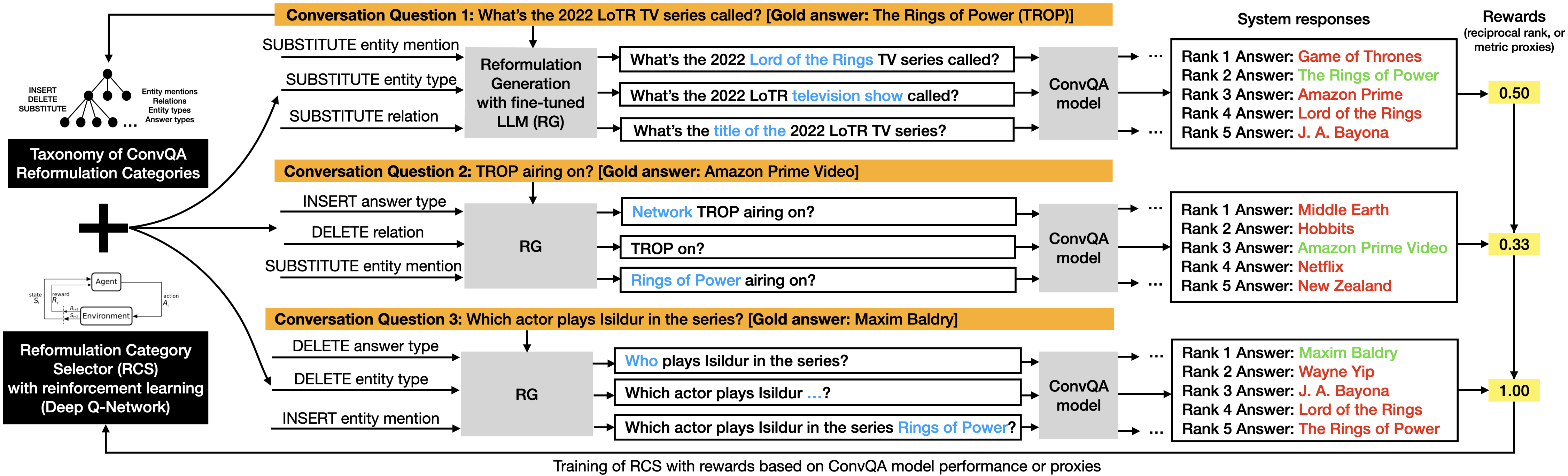}
	\vspace*{-0.7cm}
	\caption{Performance-guided reformulation generation in \reign, illustrated through our running example conversation.}
	\label{fig:runningexample}
    \vspace*{-0.1cm}
\end{figure*}

In the example conversation, $Q_4$ is phrased
in a very casual way, asking for Isildur's actor in the LOTR movie trilogy
(Peter Jackson directed the LOTR movies).
With this difficult input, the QA system may give a wrong response.
A seemingly natural approach to counter such effects would be to
have the QA system automatically reformulate the question into a more complete version~\cite{buck2018ask,vakulenko2021question,raposo2022question,yu2020few,chen2022reinforced,anantha2020open}, such as \utterance{Which actor played the role of Isildur in the Lord of the Rings movie trilogy directed by Peter Jackson?} --
this kind of run-time question rewriting to a complete natural language form in a deployed system may sometimes work, but adds inference-time overhead and may not improve performance~\cite{ishii2022can}.

\myparagraph{Approach} 
We take a different route: instead of reformulating a conversational question at inference time, we strengthen the {\em training} of the ConvQA model by exposing it upfront to a larger variety of intent-preserving surface forms for the same training sample.
Examples of such syntactic variations representing the same question semantics are in Fig.~\ref{fig:runningexample}, for $Q_1 - Q_3$ (original questions in \textcolor{orange}{orange} boxes, perturbed zones in reformulations in \textcolor{blue}{blue}).
With this more diverse training data, the ConvQA model learns to cope better with different syntactic formulations. 

Our reformulations
are created from first principles.
We propose a \textit{taxonomy} of reformulation categories for ConvQA,
that systematically manipulates parts of a given conversational question
based on string edit operations. 
For each category, we 
generate
noisy
supervision data
to fine-tune an LLM,
that then serves as our reformulation generator (RG, \textcolor{gray}{gray} boxes).
New lexico-syntactic forms in reformulations originate via use of a rich set of aliases in KGs, and world knowledge in LLMs.

Given that our generated instances are noisy, it is unlikely that for a given question, all categories of reformulations would improve the ConvQA model's performance. As a result, for each question, we would like to judiciously select \textit{a few of these} that are most beneficial.
So we pass generated reformulations to the QA model we wish to improve, and obtain ranked answer lists as responses -- shown in boxes with \textcolor{green}{green} (correct) and \textcolor{red}{red} (incorrect) answers in the right half of Fig.~\ref{fig:runningexample}.
The model's answer performance \textit{metrics} (or proxies) are used as rewards (yellow boxes)
to train a Reformulation Category Selector (RCS) with Deep Q-Networks~\cite{mnih2015human}, a form of RL that approximates \textit{value functions}. 
The trained RCS is then used as a means for \textit{model-specific data augmentation}: it selects only the top-$k$ reformulations that would be used for additional training data for the QA model for maximum performance improvement.
Instances of such question-specific categories are in Fig.~\ref{fig:runningexample} (left half).
This entire framework, termed \reign,
(\underline{REI}nforced reformulation \underline{G}e\underline{N}eration) 
is the 
main contribution of this work.
\myparagraph{Evaluation}
To assess the benefits of \reign, we perform experiments against two state-of-the-art baselines: \conquer~\cite{kaiser2021reinforcement} based on reinforcement learning, and \explaignn~\cite{christmann2023explainable} 
based on graph neural networks. Note that \reign operates by model-aware training on top of these baselines.
For test data, we leverage the generative ability of ChatGPT 
(GPT-3.5) as a proxy to obtain human-like reformulations at scale: each original question is augmented with 20 distinct reformulations.

\myparagraph{Contributions} This work calls for more robust training and evaluation of ConvQA models, our salient contributions being: 
\squishlist
    \item A novel taxonomy of question reformulations for ConvQA over KGs, based on string edit distance;
    \item A reinforcement learning model with Deep Q-Networks, that selects helpful reformulations of conversational questions guided towards better QA performance;
    \item About $335$k conversational question reformulations of test cases in two ConvQA benchmarks, suitable for rigorous evaluation of future models;
    \item The \reign framework with reusable components that judiciously augments benchmark training tailored to specific ConvQA models. All code and data are available via the project website at \url{https://reign.mpi-inf.mpg.de}.
\squishend

%% file: sections/02-concepts.tex
\section{Concepts and Notation} 
\label{sec:concepts}

\begin{table} [t]
	\centering
		\begin{tabular}{l l}
            \toprule
			\textbf{Notation}				                 & \textbf{Concept}						                    \\ \toprule
            $C, t \in \{1, 2, \ldots\}$                      & Conversation, conversational turn                        \\
            $Q=\langle q_1...q_n \rangle, A$				 & Question and its tokens, Answer  						\\	
            $Q_t, A_t$       								 & Question and answer at turn $t$                  	    \\
            $\{Q_t^i\}$                                      & Reformulations of question $Q_t$                         \\ \midrule
            $\{RC_t^i\}$                                     & RCS-predicted reformulation categories for $Q_t$         \\
            $s \in S$                                        & RCS states                                               \\
            $a \in \mathcal{A}$                              & RCS actions                                              \\
            $\mathcal{R}$                                    & RCS reward for  $Q_t^i$                                   \\
            $\Phi(\langle q_1...q_n \rangle)$                & Function to map $\langle q_i \rangle$ to state space     \\
            $M(s, \mathcal{A})$                              & Action masking vector                                    \\
            $\mathcal{Q}(s,a)$                               & Q-value (expected reward) for $a$ in $s$                 \\          
            $\mathcal{Q}^*(s,a)$                             & Optimal Q-value                                          \\
            $\pi$                                            & RCS policy                                               \\
            $\alpha$                                         & Step size in Q-Learning                                  \\
            $\gamma$                                         & Discount factor in Q-Learning                            \\ 
            $\mathbf{W_1, W_2}$                              & Weight matrices in RCS Deep Q-Network                    \\    
            $h$                                              & Hidden vector size                                       \\ 
            $d$                                              & Dimensionality of input encoding vector                  \\ 
            $Prob(\cdot)$                                    & Probability                                              \\ 
            $\tau$                                           & Boltzmann temperature for action sampling                \\ \bottomrule
    	\end{tabular}
	\caption{Notation for concepts in \reign.} 
	\label{tab:notation}
    \vspace*{-1cm}
\end{table}


Salient notation is in Table~\ref{tab:notation} (some concepts introduced in Sec.~\ref{sec:method}). 

\myparagraph{Knowledge graph} A knowledge graph (KG) consists of a set of real-world objective facts. Examples of large curated KGs (equivalently, knowledge bases or KBs) include Wikidata~\cite{vrandevcic2014wikidata}, DBpedia~\cite{auer2007dbpedia}, YAGO~\cite{suchanek2007yago}, 
or industrial ones (e.g., Google KG).

\myparagraph{Fact} A KG fact is an SPO (subject, predicate, object) triple, where a subject is an \textbf{entity} (\struct{Lord of the Rings});
an object is another entity (\struct{Maxim Baldry}), a \textbf{type} (\textit{TV Series}), or a \textbf{literal} (\struct{01 September 2022});
and a \textbf{predicate} is a relationship (\struct{cast member}) between the subject and the object.
Compound facts involving more than two entities or literals are stored as a main triple and additional $\langle$predicate, object$\rangle$ pairs (referred to as ``qualifiers'' in Wikidata~\cite{vrandevcic2014wikidata}).
For example, the main triple $\langle$\textit{The Rings of Power}, \textit{cast member}, \textit{Maxim Baldry}$\rangle$ has a qualifier $\langle$\textit{character role}, \textit{Isildur}$\rangle$. 

\myparagraph{Conversation} A conversation $C$ consists of a sequence of $\langle Q_t, A_t \rangle$ \textit{turns} around a topic of interest. An example is in Sec.~\ref{sec:intro}.

\myparagraph{Intent} An intent is a specific information need: a conversational question and its reformulations share the same intent. In this work, each training and test question in a benchmark represents a unique intent: the reformulations of the training and test cases have different surface forms while preserving the original intent.

\myparagraph{Question} A question $Q$ manifests an intent and consists of a sequence of tokens $\langle q_1...q_n \rangle$. $Q$ can either be complete (explicit expression of intent), like \utterance{What's the 2022 LOTR TV series called?} ($Q_1$), or incomplete (implicit expression of intent), like \utterance{And who in the Jackson trilogy?} ($Q_4$).

\myparagraph{Answer} An answer $A$ is a response to the information need in question $Q$ (\struct{Harry Sinclair} is the answer $A_4$ to $Q_4$). 
In this work, an answer can be a KG entity, a type, or a literal. 
It can either be a ConvQA model's response or a gold answer from a benchmark.

\myparagraph{Reformulation} A question reformulation is obtained by transforming a question into a different surface form with the same intent.
A reformulation is generated using an $\langle operation, operand \rangle$ pair and the original question. Here, \textit{operations} could be \{insertion, deletion, substitution\}, while \textit{operands} could be \{entities, predicates, question entity types, expected answer types\}. An example transformation is adding an answer type to question $Q_2$: \utterance{TROP airing on?}, to produce the reformulation $Q_2^1$: \utterance{Network TROP airing on?}.

\myparagraph{Mention} A mention
refers to a sequence of tokens in $Q$ that is the surface form of a KG item (entity, predicate, or type). A mention of a predicate is
referred to as a relation. 
For example, in
$Q_2^1$: \utterance{Network TROP airing on?}, \phrase{Network}, \phrase{TROP}, and \phrase{airing on} are mentions of 
KG answer type \struct{video streaming service}, KG entity \struct{The Rings of Power (TROP)}, and KG predicate \struct{original broadcaster}, respectively. 




%% file: sections/03-method.tex
\section{The \reign framework}
\label{sec:method}

\begin{figure} [t]
	\centering
	\includegraphics[width=\columnwidth]{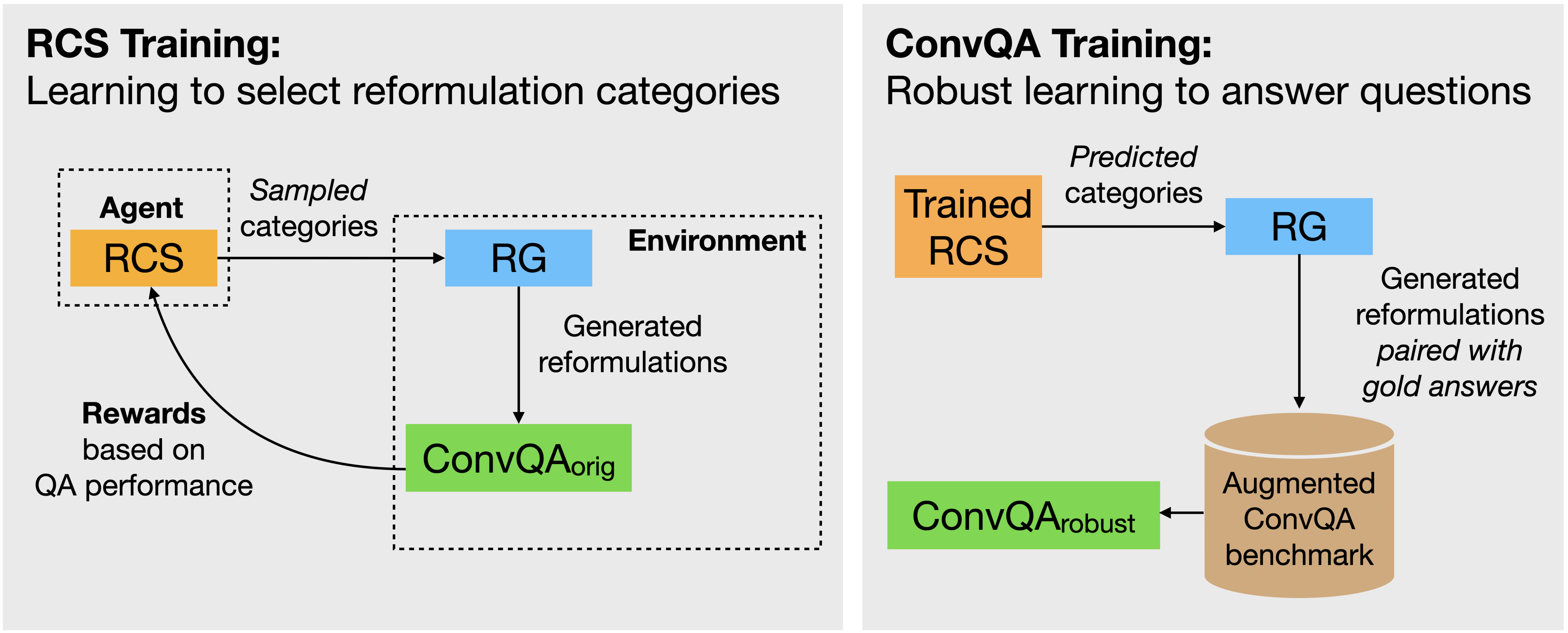}
    \vspace*{-0.7cm}
	\caption{Workflow of \reign: RCS is trained by reinforcement learning, and RG by supervised learning.} 
	\label{fig:overview}
    \vspace*{-0.5cm}
\end{figure}

An overview of the workflow in the proposed \reign architecture is depicted in Fig.~\ref{fig:overview}. The pipeline consists of three trainable models, where the first two are our contributions: 
\squishlist
\item A {\bf reformulation category selector (RCS) model}, that takes a question $Q_t$ as input, and produces a reformulation category $RC_t^i$ for transforming $Q_t$, as output;
\item A {\bf reformulation generator (RG) model}, that takes some $Q_t$ and $RC_t^i$ as input, and produces a reformulation $Q_t^i$ of $Q$ according to $RC^i_t$, as output;
\item An {\bf external ConvQA model}, that takes some $Q_t$ as input, and produces a ranked list of answers $\langle A_t\rangle$ as output.
\squishend

Reinforcement learning (RL) is used to train the RCS model (Deep Q-Networks~\cite{mnih2015human} in this work), with the goal of learning to select the most suitable transformation categories given a specific question, using existing QA performance metrics or suitable alternatives as reward signals. 
The categories come from our novel reformulation taxonomy. The RG model is trained with (distantly) supervised learning (SL), using an LLM (BART in our case~\cite{lewis2020bart}) fine-tuned with questions paired with a specific category and the resulting reformulation in the form $\langle(Q_t, RC_t^i);Q_t^i\rangle$. 
This is distant supervision in the sense that the reformulations used for fine-tuning are generated in a noisy manner using rules following our taxonomy, and are not human reformulations. 
The ConvQA model used could be trained with SL~\cite{christmann2023explainable,kacupaj2022contrastive,shen2019multi} or RL~\cite{kaiser2021reinforcement}, according to its original training paradigm. In Fig.~\ref{fig:overview}, the original model $\mathrm{ConvQA_{orig}}$ is trained with $\langle Q_t, A_t\rangle$ pairs in a ConvQA benchmark, while the more robust model $\mathrm{ConvQA_{robust}}$  is trained on additional QA pairs where the reformulations $\{Q_t^i\}$ for a specific $Q_t$ are also paired with the original gold answer $A_t$. We now describe each component. 

\section{Reformulation Category Selector} 
\label{sec:rcs}

\subsection{Reformulation taxonomy}
\label{subsec:taxonomy}

\begin{figure} [t]
	\centering
	\includegraphics[width=\columnwidth]{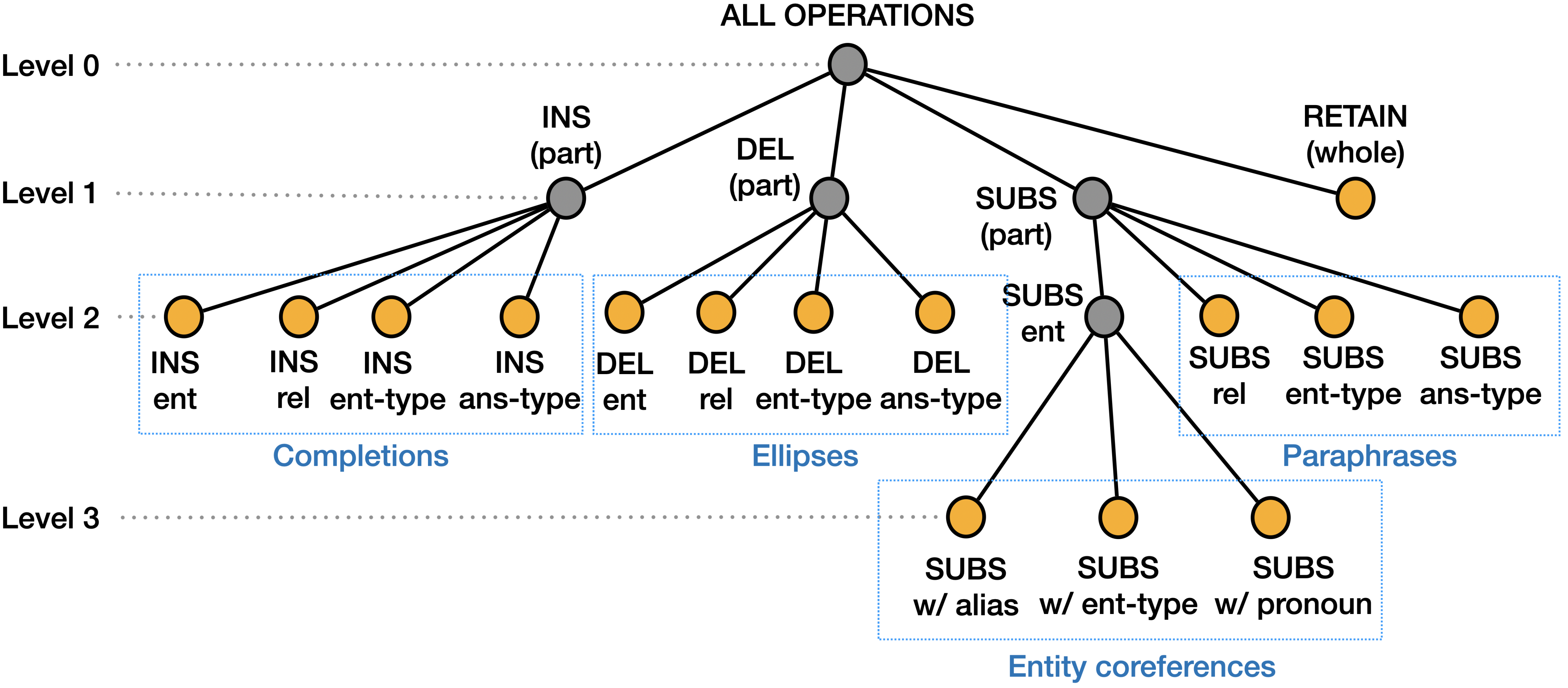}
    \vspace*{-0.7cm}	
    \caption{Taxonomy of reformulation categories. Legend: part = question-part; INS = Insert, DEL = Delete, SUBS = Substitute; ent = entity mention, rel = relation, ent-type = entity type mention, ans-type = answer type mention; w/ = with.}
	\label{fig:taxonomy}
    \vspace*{-0.55cm}
\end{figure}

\myparagraph{Categories} We propose a taxonomy of reformulations, a topic that has mostly been treated as monolithic in past work~\cite{buck2018ask,hermjakob2002natural,kaiser2021reinforcement}.
To begin with, observe that a reformulation of a conversational question is a \textit{modification} of its basic \textit{parts}.
Thus, a systematic generation of reformulations involves an understanding of these parts and meaningful modifications.
For (Conv)QA over KGs, these basic question components comprise mentions of one or more entities, their types, predicates, and expected answer types.
In analogy with string edit operations, our modifications include insertion, deletion and substitution. Transposition could be another basic operation, but we do not consider that in this work as reordering question phrases has little effect on several retrieval models.
Viewing these three operations and the four parts of a question as operands, we 
obtain a taxonomy as shown in Fig.~\ref{fig:taxonomy},
where reformulation categories are leaf nodes (marked \textcolor{orange}{orange}). Examples are \phrase{INSERT entity-type}, \phrase{SUBSTITUTE relation}, and \phrase{DELETE relation}. 
Note that we require our reformulations to be \textit{intent-preserving}: this imposes constraints on what we can insert or substitute in the original question.
We cannot, for example, replace an entity or relation by a different one -- that would disturb the semantics of the conversation as a whole.

\myparagraph{Phenomena} As shown with dashed boxes in Fig.~\ref{fig:taxonomy}, our taxonomy subsumes several classes of conversational phenomena:
\squishlist
\item Insertions \textit{complete} the question to a more intent-explicit form;
\item Deletions cause \textit{ellipses} in context;
\item Substitutions create \textit{paraphrases}; 
\item Substituting entity mentions specifically leads to \textit{coreferencing}.
\squishend
The last operation can be sub-divided into three categories as per the case of substitution with a pronoun (\phrase{TROP} $\mapsto$ \phrase{it}) or with its type (\phrase{TROP} $\mapsto$ \phrase{the series}) or with an alias (\phrase{TROP} $\mapsto$ \phrase{Rings of Power}). 
A special case in our taxonomy is the operation \phrase{RETAIN whole question}, where the question is left as such: it can be considered as a degenerate reformulation.
Finally, we have $\mathbf{15}$ \textbf{reformulation categories}, corresponding to these leaf nodes.

\subsection{Training the RCS model}
\label{subsec:train-rcs}

\myparagraph{Overall idea} Given an input question and the taxonomy, we would like the Reformulation Category Selector (RCS) model to suggest some categories so that training with reformulations that belong to these categories would lead to better QA performance \textit{metrics}. This, in turn, means that we would like the RCS to estimate \textit{values} that correspond as much as possible to such metrics.
This motivates us to use Deep Q-Networks (DQN)~\cite{mnih2015human}, a reinforcement learning approach that directly learns a value function approximator using QA metrics as rewards. The estimate of the value of a (state, action) pair is in turn used to infer the policy for predicting or sampling actions. This is in contrast to the relatively more popular choice of learning policy gradients that directly model action probabilities given an input state (for example, the REINFORCE algorithm~\cite{williams1992simple}).

Concretely, we employ DQNs to train an agent (the RCS model) to select actions $(a \in \mathcal{A})$ (reformulation categories) given a current state $(s \in S)$ (the input question $Q_t$). The agent interacts with an environment (the reformulation generator and the ConvQA model, described later).
This environment provides the next state $s' \in S$ (the generated reformulation $Q_t^i$) and a reward $\mathcal{R}$ (QA performance metric) to the agent. This is a Markov Decision Process (MDP) comprising states, actions, the transition function, and rewards $(\mathcal{S}, \mathcal{A}, \delta, \mathcal{R})$, where the individual parts are defined next. 
Algorithm~\ref{algo:reign} shows the precise steps of applying Deep Q-Learning in the RCS model. 

\myparagraph{States} A state $s \in \mathcal{S}$ is defined by a conversational question, represented by its encoding with function $\Phi$: $\boldsymbol{s}= \Phi(Q)$ (lines 3, 7 in Algorithm~\ref{algo:reign}; BERT~\cite{devlin2018bert} embeddings in our experiments averaged over each question token, and over all hidden layers). 

\myparagraph{Actions} The set of actions $\mathcal{A}$ corresponds to the $15$ reformulation categories from our taxonomy. 
Note that every action (category) may not be available at every state. For instance, when a question does not have any mention of an entity type, it is not meaningful to apply the actions of deletion or substitution of an entity type. 
Therefore, we use \textit{action masking} as follows to allow only \textit{valid actions} to be chosen given the current state (when this information is available).
A masking vector $M(s, \mathcal{A})$ that has ones at indices corresponding to valid actions, and zeros elsewhere, is element-wise multiplied with the vector containing learnt probabilities of actions at a given state.

\myparagraph{Transitions} The transition function $\delta$ is deterministic and updates the state $s \in S$ by applying one of the actions $a \in \mathcal{A}$, resulting in a new state $s' \in S$ that corresponds to the encoding of the reformulation $Q^i$. The resulting reformulation is obtained by invoking the RG model (line 6 in Algorithm~\ref{algo:reign}).
 
\myparagraph{Rewards} The reward $\mathcal{R}$ models the quality of the chosen action, and guides the agent towards its goal, which here is an improved answering performance. 
When a selected category leads to a reformulation on which the ConvQA model obtains \textit{better performance than the original question}, the agent should get a high reward, and vice versa (line 8 calls the ConvQA model for this reward). 
Thus, an obvious choice here is to use any desired QA performance metric as the reward.
We use the reciprocal rank (RR)~\cite{voorhees1999trec} metric in this work, that is the reciprocal of the first rank at which a gold answer is found. We use it for the following reasons: (i) we have binary relevance of response entities, either correct ($1$) or incorrect ($0$); (ii) we deal with factoid QA, and there are usually only a few correct answers (typically between one and three for the benchmarks used). Formally, this reward based on \textit{reciprocal rank difference} is computed as:
\begin{equation}
   \mathcal{R} = \mathrm{ReciprocalRank}(\langle A_t^i \rangle) - \mathrm{ReciprocalRank}(\langle A_t \rangle) \\ 
\end{equation}
where $\langle A_t \rangle$ and $\langle A_t^i \rangle$ are the ranked lists of responses by the ConvQA model to $Q_t$ and $Q_t^i$, respectively.
If the correct answer was not found in the top few positions (five in our experiments), then we set the reciprocal rank value to $-1$.
The range of this reward then lies in the closed interval $[-2, +2]$.
Since this nicely corresponds to a symmetric positive reward and punishment in respective cases of success and failure, we do not perform any further reward normalization. Note, however, that our framework can be used with any metric of choice: the use of RL removes the dependency on the metric being differentiable.

\myparagraph{Algorithm} As motivated before, we use Deep Q-Networks (DQN) as our RL algorithm, which is a \textit{model-free, value-based method} that learns to predict so-called Q-values $\mathcal{Q}(s,a)$ for each state-action pair to quantify the usefulness of taking action $a$ in state $s$ under a policy $\pi$~\cite{mnih2015human}. 
The policy is a function mapping states to actions based on the Q-values.
The main update step in Q-Learning is: 
\begin{equation}
    \mathcal{Q}(s,a) \leftarrow  \mathcal{Q}(s,a) + \alpha \cdot  [ (\mathcal{R} + \gamma \cdot \max_{a'}  \mathcal{Q}(s',a')) -  \mathcal{Q}(s,a)]
    \label{eq:qlearning}
\end{equation}
where $\alpha$ is the step size and $\gamma$ is the discount factor that determines how much influence the next state's estimate has on the current state. The expression $(\mathcal{R} + \gamma \cdot \max_{a'}  \mathcal{Q}(s',a'))$ is called TD (temporal difference) target, and the term inside square brackets $[\ldots]$ is called TD error. 
$\mathcal{Q}(s,a)$ is randomly initialized, except for terminal states, where this is zero. 
In practice, the parameters of the DQN are updated batch-wise (lines $10-18$).
A batch consists of a set of \textit{experiences}: each experience is a tuple of the form $(s, a, s', r)$ (line 9).
The updated parameters $\theta$ are obtained by calculating the Mean-Squared Error between the TD target and the current Q-values of each state-action pair in the batch (line 19). 
The objective function is to maximize the expected reward.
Since our state space is large, we cannot directly
learn 
tabular entries
for each $\langle s,a \rangle$ pair, as was typical in more traditional RL setups. 
Instead, Q-values are predicted via a neural Q-network
with trainable parameters $\theta$ (a two-layer feed-forward network in our case): 
\begin{equation}
    \mathcal{Q}_{\theta}(s,\langle a \rangle) = M(s, \mathcal{A}) \circ (\mathbf{W_2} \times \mathrm{ReLU}(\mathbf{W_1} \times \boldsymbol{s}))
\end{equation}
where 
$\mathcal{Q}_{\theta}(s, \langle a \rangle)$ is a function returning a vector of size $|\mathcal{A}|$ and stores the obtained values for every action $a \in \mathcal{A}$ given some $s \in \mathcal{S}$; $\boldsymbol{s} \in \mathbb{R}^{d \times 1}$;
$d$ is the size of the input encoding vector;
$\mathbf{W_1} \in \mathbb{R}^{h \times d}$, $\mathbf{W_2} \in \mathbb{R}^{|\mathcal{A}| \times h}$ are the weight matrices;
$M(s, \mathcal{A}) \in \mathbb{R}^{|\mathcal{A}| \times h}$ is the action mask; and hidden size $h$ is a tunable hyperparameter. 
ReLU is the non-linear activation function. 

During training, the agent needs to explore different actions in each state via sampling. In this work, we sample from a Boltzmann distribution to enable such exploration (line 5). A Boltzmann distribution is parameterized by a temperature $\tau$ that we can use to conveniently control the degree of exploration:
\begin{equation}
    \mathrm{Prob}(a^{sample},  \tau) = \frac{e^{\mathcal{Q}_\theta(s,a^{sample})/\tau}}{\sum\limits_{a \in \mathcal{A}} e^{\mathcal{Q}_\theta(s,a)/\tau}}
    \label{eq:boltzmann}
\end{equation}
A $\tau$-value close to zero means taking the best action (with highest reward at this point) greedily more often, whereas larger values ($\tau$ is unbounded) make the actual Q-values less relevant and result in a random policy. 

\begin{algorithm} [t]
	\small
	\DontPrintSemicolon
	\SetAlgoLined
	\SetKwInput{Input}{Input}\SetKwInput{Output}{Output}
	\SetKwComment{Comment}{$\blacktriangleright$\ }{}
		\Input{Sequence of conversational questions $\langle Q \rangle$,  step size ($\alpha > 0$), discount factor ($\gamma > 0$), Boltzmann temperature ($\tau > 0$),\\size of update ($batchSize$), initial DQN parameters $\boldsymbol{\theta}$} 
		\Output{Updated DQN parameters $\boldsymbol{\theta}$}
	$experience \gets \langle \rangle$ \Comment{Initialize experience queue}
	\ForEach{$Q_t \in \langle Q \rangle$}
    	{$\boldsymbol{s} \gets \Phi(Q_t)$ \Comment{Encode question}
            $\mathcal{Q}_{\theta}(s,\langle a \rangle) = M(s, \mathcal{A}) \circ (\mathbf{W_2} \times \mathrm{ReLU}(\mathbf{W_1} \times \boldsymbol{s}))$ \Comment{Get Q-values for actions in s}
            $a^{sample} \sim  \textrm{Prob}(\mathcal{A}, \tau)$ \Comment{Use Eq.~\ref{eq:boltzmann}} 
            $Q_t^{a^{sample}} \gets  \textrm{RG}(\langle Q_1 \ldots A_{t-1} Q_t RC_t^{a_{sample}}\rangle)$ \Comment{Invoke RG}
            $\boldsymbol{s'} \gets \Phi(Q_t^{a_{sample}})$ \Comment{Encode reformulation}
            $r \gets \mathcal{R}(\textrm{ConvQA}(Q_t))$ \Comment{Invoke ConvQA for reward}
            $experience.\textrm{enqueue}(s,a^{sample}, s',r)$ \Comment{Store experience}    
   
	\uIf{$|experience| >= batchSize$}
	 {
    $batch \gets experience.\textrm{dequeue}(batchSize)$\;
   $q^{batch} \gets \langle \rangle$ \Comment{Initialize queue for 
   Q-values}
   $q^{targets} \gets \langle \rangle$ \Comment{Initialize queue for 
   TD targets}
    \ForEach {$(s,a,s', r) \in batch$}
           	{$q^{batch}.\textrm{enqueue}(\mathcal{Q}_{\theta}(s,a))$ \;
            $q^{targets}.\textrm{enqueue}(r + \gamma \cdot \max_{a'}  \mathcal{Q}_{\theta}(s',a')) $\;
    }
    $\boldsymbol{\theta} \gets \boldsymbol{\theta} - \alpha \nabla  1/batchSize \cdot \sum_{i=0}^{batchSize} (q^{targets}_{i}-q^{batch}_{i})^2$ 
    } }
    \Return{$\boldsymbol{\theta}$}
\caption{Deep Q-Learning in RCS model}
\label{algo:reign}
\setlength{\textfloatsep}{0pt} 
\end{algorithm}
\subsection{Applying the RCS model}
\label{subsec:rcs-test}

We train the RCS on the development set\footnote{In this work, we reuse ConvQA benchmark development sets for training the RCS model, fine-tuning the RG model, adjusting hyperparameters for all models, and selecting best \reign configurations. We intentionally avoid large-scale learning of RCS and RG on train sets to stress-test generalizability of \reign components: only the QA model is learnt from the full train set. Further, any kind of leakage to test sets is thereby precluded for all models.} of a ConvQA benchmark, and apply it on the questions in the training set.
At \textit{RCS inference time}, the agent follows a greedy policy $\pi$ with respect to Q-values, and typically chooses an action $a^{greedy}$ 
in a state $s$ as below:
\begin{equation}
    a^{greedy} = \pi_s = \arg \max_{a \in \mathcal{A}} \mathcal{Q}(s,a)
\end{equation}
In our case, we take the top-$k$ predicted reformulation categories 
$\{RC_t^{1}, \ldots, RC_t^{k}\}$
from the RCS (instead of the top-1 from the $\arg \max$) for each training question $Q_t$. The RG then picks these up to actually generate the new question variants $\{Q_t^1, \ldots, Q_t^k\}$. 

\section{Reformulation Generator}
\label{sec:rg}

\subsection{Training the RG model}
\label{subsec:rg-train}

\myparagraph{Basic setup} The reformulation generator (RG) is implemented by fine-tuning a pre-trained LLM for sequence generation (BART~\cite{lewis2020bart} in our case). 
BART is especially effective when information is both copied from the input plus perturbed with noise, to generate the output autoregressively~\cite{lewis2020bart}: this is exactly the setup in this work.
The concatenation of the conversation history $\langle Q_1 A_1 \ldots Q_{t-1} A_{t-1}\rangle$, the current question $Q_t$, and a special reformulation category tag (\phrase{rc1}, \phrase{rc2}, ...\phrase{rc15}) constitute the input,
and the category-specific reformulation is the output.

\myparagraph{Noisy data for fine-tuning} We generate
fine-tuning data for the BART model through \textit{distant supervision}.
This is a noisy process, but the alternative of strong supervision would entail the use of human-generated reformulations.
This would be expensive to obtain at scale (benchmarks like \textsc{ConvRef}~\cite{kaiser2021reinforcement} contain a relatively small number of unique reformulations, and lack category labels). 
Further, given a conversational question, an average crowdworker is not likely to be able to come up with several diverse and distinct reformulations for each category.

Specifically, we adopt the following strategy.
First, we need to find mentions of entities, types, and predicates in the question.
We use \textsc{Tagme}~\cite{ferragina2010tagme} for entity
mention
detection and the Wikidata Search API\footnote{\url{https://www.mediawiki.org/wiki/API:Main_page}} for linking them to the KG. 
The recently proposed \textsc{Clocq} API~\cite{christmann2022beyond} is used to obtain type and alias information. 
We use aliases in curated KGs
for synonyms of entities and predicates, a rich and precise yet relatively under-explored resource. 
We annotate entity type mentions by searching the question before and after any entity mention in the case of question types, and after question words in the case of answer types. 
To obtain predicate annotations, we extract KG paths connecting the linked question entities and answer entities,
or between question entities (e.g. in case of yes/no questions) using the \textsc{Clocq} API. The similarity of each path with respect to the question is scored with SBERT~\cite{reimers2019sentence} and the predicate on the top-scoring path is assumed to be the intended predicate in the question. We only keep the top 100 predicates found this way to reduce noise. Predicate mentions (relations) in the question are extracted by searching for verbs, the predicate label in the KG, or its alias(es) in the question. If not found, we remove mentions of the entities and types, and question words from the question, and treat the remainder as the predicate mention.

Once we have mentions and their disambiguations, we can apply our transformations from the taxonomy (Fig.~\ref{fig:taxonomy}) on input questions. Deletion is straightforward: the mention is simply removed from the question token sequence. For substitution, the main decision to make is the source of alternative surface forms for the linked KG items.
Substitution happens in-place: the source mention is replaced by the target mention from the KG alias list in its corresponding position in the question.
Each unique alias results in a unique transformation possibility.
Pronoun replacements for human entities are performed by looking up their gender in the KG.
For insertions, the main concern is the position of insertion in the question: (i) mentions of answer types and relations are inserted just after the \textit{wh-} question word; (ii) mentions of entity types are inserted just before the respective entity; and (iii) entity mentions 
are inserted at the end of the question.
The strategy above entails that some categories will have an extremely large number of training cases (like substitutions, as we very often have $5+$ synonyms for an entity or predicate) while others would have relatively lower volumes (like relation deletions). 
To increase the number of data points for sparse categories, we adopt a \textit{back-and-forth strategy}: it is easy to add an entity
mention 
to a question when none are detected (like \utterance{Directed by...?} $\mapsto$ \utterance{LOTR part 1 directed by?}), and then the \textit{reverse} of this rule would give us a sample of entity deletion.

\subsection{Applying the RG model}
\label{subsec:rg-test}

The BART model is fine-tuned on distantly supervised data generated with the ConvQA dev set.
It is then applied on the train set where a question and a category from the RCS are already available.

\vspace{-0.2cm}
\section{Conversational question answering}
\label{sec:convqa}

\subsection{Training the ConvQA model}
\label{subsec:convqa-train}

A ConvQA model is trained on sequences of $\langle Q_t, A_t \rangle$ pairs. In the original training mode, QA pairs are directly used from the benchmark train sets. This original or initial QA model is used to collect rewards for the RCS in one pass over the dev set (as mentioned earlier, the QA dev set is used to train the RCS model). After the trained RCS and RG models generate the reformulations for each training question, these reformulations are paired with the corresponding gold answer of the original training question. These new $\langle$reformulation, gold answer$\rangle$ pairs are added to the benchmark, and the ConvQA model is trained again on this augmented resource. This model is expected to be more robust than the original model (original and robust models are marked ConvQA$_\mathrm{orig}$ and ConvQA$_\mathrm{robust}$ in Fig.~\ref{fig:overview}, respectively).

\subsection{Applying the ConvQA model}
\label{subsec:convqa-test}

The trained ConvQA model is directly applied to the questions
in test sets at \textit{answering time} to produce ranked lists of entities. 

%% file: sections/04-experiments.tex
\section{Experimental Setup}
\label{sec:setup}

\myparagraph{Benchmarks} As shown in Table~\ref{tab:data}, we use two ConvQA benchmarks: \convmix~\cite{christmann2022conversational} (more recent) and \convquestions~\cite{christmann2019look} (more popular). These contain realistic questions from crowdworkers. We obtained $20$ reformulations from ChatGPT (\texttt{gpt-3.5-turbo} model) for each test question in these benchmarks,
 with ten each from two different settings: (i) one asked for reformulations with access to the full conversation history (previous questions and gold answers), and (ii) the other only with the current question.
 Examples are in Table~\ref{tab:gpt-reformulations}. All GPT-generated reformulations are available at our website \url{https://reign.mpi-inf.mpg.de}.
For prompting GPT, we tried a few alternatives.
We saw that examples did not have a noticeable effect on the generations. Thus, we used the following zero-shot prompt (the `History' line is omitted for generating the variants without history) and set the temperature value to zero, for obtaining deterministic behavior (as far as possible): 
\begin{Snugshade}
\noindent \textit{Reformulate the `Question' 10 times in a short, informal way. Assume third person singular if not obvious from the question.\\
`History': \textsc{\{Conversation history\}}\\
`Question':  \textsc{\{Question\}}  \\
`Reformulation': }
\end{Snugshade}
The second sentence was used to avoid generations like \utterance{Your place of birth?} instead of the correct \utterance{His ...?} or \utterance{Her ...?} There are \textit{no duplicates} in any of the ChatGPT reformulations. Conversations in \convquestions are generated by permuting questions from a seed set of $700$ conversations: we used only the train set for this seed ($420$ conversations) for training ConvQA models, to decouple the effect of data augmentation inherent in the benchmark.

\begin{table} [t] 
	\newcolumntype{G}{>{\columncolor [gray] {0.90}}c}
	\resizebox{\columnwidth}{!}{
	\begin{tabular}{G c c c G} 
		\toprule
		   \textbf{Benchmark} &	\textbf{Train}  &	\textbf{Dev}	& 	\textbf{Test}	& \textbf{GPT-Test} \\ \toprule 
            \convmix~\cite{christmann2022conversational}    & 8.4k (1680) & 2.8k (560) & 4.8k (760)  & 100.8k (760) \\
            \convquestions~\cite{christmann2019look} & 33.6k (6720) & 11.2k (2240) & 11.2k (2240) & 235.2k (2240) \\  \bottomrule
	\end{tabular} }
	\caption{Benchmark sizes as \#questions (\#conversations). Reformulations are also counted as individual questions to be answered. Questions for the GPT-Test sets subsume the original test questions.}
	\label{tab:data}
    \vspace*{-0.9cm}
\end{table}

\begin{table} [t] 
	\centering
		\begin{tabular}{p{8cm}}
  \toprule
          \textbf{[Books]} \textbf{History:}  \utterance{How many Pulitzer Prizes has John Updike won? 2.} \\ \textbf{Question: } \utterance{Which was the first book to win him the award?}     \\ \hdashline
              \textbf{Ref 1:} \utterance{What book earned John Updike his first Pulitzer Prize?}  \\
            \textbf{Ref 2:}   \utterance{What was the author's first book to win a Pulitzer?}        \\
        \textbf{Ref 3: } \utterance{Title of John Updike's first Pulitzer Prize-winning book?}  \\ \midrule
       
        \textbf{[Movies]} \textbf{History:} \utterance{ Which year did the Hobbit An unexpected journey released? 2012. } \\ \textbf{Question: } \utterance{What is the book based on?}   \\ \hdashline
             \textbf{Ref 1:} \utterance{What's the book about?}  \\
           \textbf{Ref 2:}   \utterance{ What's the book's topic?}         \\
    	   \textbf{Ref 3:}  \utterance{What's the book's subject?}   \\ \midrule
    
    \textbf{[Music]} \textbf{History:} \utterance{ Which singer sang the number Single Ladies? Beyonce.  What is the year of its release? 2008.  Who is her spouse? Jay-Z .  What is his date of birth? 4 December 1969. } \\ \textbf{Question: } \utterance{Was Kanye West a composer of the song?}   \\ \hdashline
             \textbf{Ref 1:} \utterance{Did Kanye West contribute to the lyrics of the song?}  \\
           \textbf{Ref 2:}   \utterance{ Did Kanye West perform the song with Beyonce?}         \\
    	   \textbf{Ref 3:}  \utterance{Was Kanye West featured in the song?}   \\ \midrule

        \textbf{[TV series]} \textbf{History:} \utterance{What is the release year of the TV series See? 2019.} \\ \textbf{Question: } \utterance{created by?}   \\ \hdashline
             \textbf{Ref 1:} \utterance{Who's responsible for it?}  \\
           \textbf{Ref 2:}   \utterance{ Who's the mastermind?}         \\
    	   \textbf{Ref 3:}  \utterance{Who's the author?}   \\ \midrule
        
             \textbf{[Soccer]} \textbf{History:} \utterance{Pele scored how many goals in international play? 77.  Has he scored the most goals? No. } \\ \textbf{Question: } \utterance{Did Messi beat his goal total?}   \\ \hdashline
             \textbf{Ref 1:} \utterance{Did Messi surpass Pele's international goal record?}  \\
           \textbf{Ref 2:}   \utterance{ Has Messi scored more international goals than Pele?}         \\
    	   \textbf{Ref 3:}  \utterance{Did Messi break Pele's goal-scoring record?}   \\ 
                     \bottomrule
	\end{tabular}  
	\caption{Examples of GPT reformulations for test sets.}
	\label{tab:gpt-reformulations}
    \vspace*{-0.9cm}
\end{table}

\myparagraph{Baselines} ConvQA models belong to two families, one based on \textit{history modeling}, and the other on \textit{question completion} (Sec.~\ref{sec:related}). We choose one open-source system from each family for KG-QA: \conquer~\cite{kaiser2021reinforcement} (history modeling with context entities, with RL) and the very recent \explaignn (completion to an intent-explicit structured representation, with GNN). \explaignn was built for heterogeneous sources, and we use the KG-only model, in line with our setting. Default configurations were used for both systems.

\myparagraph{Metrics} All methods produce ranked lists of entities with binary relevance. We thus used three appropriate KG-QA metrics~\cite{saharoy2022question}: Precision@1 (\textbf{P@1}), Mean Reciprocal Rank (\textbf{MRR}), and whether a correct answer is in the top-5 (\textbf{Hit@5}). We define a new metric \textbf{Robust}, that computes, for each question, the number of reformulations correctly answerable by a ConvQA model, averaged over the number of test intents. The Robust measure lies between 0 and
the number of reformulations per question including the original
formulation (hence 21 in our case). The higher this value, the more robust the model.
Statistical significance (*) is conducted via McNemar's test for binary variables (P@1 and Hit@5), and $2$-tailed paired $t$-test otherwise (MRR, Robust), with $p < 0.05$.

\myparagraph{Initializing \reign} 
We use Wikidata as our KG: all models use the dump from \texttt{31-01-2022}. 
We use 
BART (\url{bit.ly/3N9WPVj}, for RG), and BERT (\url{bit.ly/3NkKRsd}, for state encoding in RCS) implementations
from Hugging Face.
As history input to BART, we used only the first and previous turns of the conversation~\cite{qu2019bert,vakulenko2021question}.
Hyperparameters for the Deep Q-Network in the RCS were tuned on the \convmix dev set: $d = 768$, hidden size $h = 128$, Boltzmann temperature $\tau = 0.3$, discount factor $\gamma = 1.0$ (no decay for future rewards), step size $\alpha = 10^{-5}$, batch size $= 10$, and epochs $= 5$. 
The RG was trained for $3$ epochs, and $2k$ examples from each reformulation category were used for fine-tuning BART.
Both RCS and RG models are only trained on \convmix and applied \textit{zero-shot} on \convquestions.
Five reformulation categories were selected by RCS for every question ($k=5$). 
A single GPU (NVIDIA Quadro RTX 8000, 48 GB GDDR6) was used to train and evaluate all models. The TensorFlow Agents library is used for the RL components.

%% file: sections/05-results.tex
\section{Results and Insights}
\label{sec:results}


\subsection{Key findings}
\label{subsec:key}

\begin{table} [t] 
		\begin{tabular}{p{8cm}}
			\toprule
     \textbf{[Books]} \textbf{History:} \utterance{Which book won the 2017 Pulitzer Prize for Fiction? The Underground Railroad.  subject of the book? Slavery in the United States.  publisher of the novel? Doubleday.  } \\ \textbf{Question: } \utterance{author of the fiction?}   \\ \hdashline
             \textbf{Ref 1:} \utterance{creator of the fiction?} [SUBS rel] \\
           \textbf{Ref 2:}   \utterance{ Which individual is author of the fiction?}  [INS ans-type]       \\
    	   \textbf{Ref 3:}  \utterance{author of the fiction The Underground Railroad?}  [INS ent] \\ \midrule

		 \textbf{[Movies]} \textbf{History:}  \utterance{Who was the director of The Lord of the Rings? Peter Jackson.} \textbf{Question:} \utterance{Who played Frodo Baggins?}    \\ \hdashline
               \textbf{Ref 1:}  \utterance{ Who  Frodo Baggins?}    [DEL rel]\\
            \textbf{Ref 2:}  \utterance{Who portrayed Frodo Baggins ?}       [SUBS rel]  \\ 
             \textbf{Ref 3:}  \utterance{Who played Frodo Baggins in it?}       [SUBS pronoun]   \\ 
		  
           \midrule  

         \textbf{[Music]} \textbf{History:}  \utterance{-} \\ \textbf{Question: } \utterance{Formation year of the band U2?}     \\ \hdashline
               \textbf{Ref 1:} \utterance{Formation year of the rock band U2?} [SUBS ent-type]  \\
             \textbf{Ref 2:}   \utterance{Which year is Formation year of the band U2?}  [INS ans-type]       \\
     	   \textbf{Ref 3:}  \utterance{Formation year of U2?} [DEL ent-type]     \\ \midrule
         
                       \textbf{[TV series]}  \textbf{History:} \utterance{Who played as Marty in Ozark series? Jason Bateman.  and Wendy Byrde? Laura Linney.  who is the director of the series? Jason Bateman.  How many episodes are in the series? 30. } \\ 
                       \textbf{Question: } \utterance{production company of the series?}   \\ \hdashline
             \textbf{Ref 1:} \utterance{production company of the series television series?} [INS ent-type] (noisy) \\
           \textbf{Ref 2:}   \utterance{ production company of the series Ozark?}  [INS ent]       \\
    	   \textbf{Ref 3:}  \utterance{production house of the series?}  [SUBS rel] \\ \midrule

      \textbf{[Soccer]} \textbf{History:} \utterance{What is the full name of footballer Neymar? Neymar da Silva Santos Junior.  Birthplace of Neymar? Brazil .  When was he born? 5 February 1992. } \\ \textbf{Question: } \utterance{Which club does he play now?}   \\ \hdashline
              \textbf{Ref 1:} \utterance{Which club does he play now association football player?} [INS ent-type] \\
            \textbf{Ref 2:}   \utterance{Which club does he play now Neymar?}  [INS ent]       \\
     	   \textbf{Ref 3:}  \utterance{Which Football team does he play now?}  [SUBS ans-type] \\ 
       
                     \bottomrule
	\end{tabular}  
	\caption{Examples of \reign-generated reformulations along with respective reformulation categories, used for training.}
	\label{tab:bart-reformulations}
    \vspace*{-0.8cm}
\end{table}

\begin{table*} [t] 
	\newcolumntype{G}{>{\columncolor [gray] {0.90}}c}
	\resizebox{\textwidth}{!}{
	\begin{tabular}{l G G G c c c c  G G G c c c c}
		\toprule
		  \textbf{Benchmark} $\rightarrow$ &	\multicolumn{3}{G}{\textbf{\convmix~\cite{christmann2022conversational} Test}}  & \multicolumn{4}{c}{\textbf{GPT-\convmix Test}} &  \multicolumn{3}{G}{\textbf{\convquestions~\cite{christmann2019look} Test}} & \multicolumn{4}{c}{\textbf{GPT-\convquestions Test}} \\ \midrule
       \textbf{Method} $\downarrow$  &  \textbf{P@1} &	\textbf{MRR}	& 	\textbf{Hit@5}	&	 \textbf{P@1} &	\textbf{MRR}	& 	\textbf{Hit@5}	& \textbf{Robust} &   \textbf{P@1} &	\textbf{MRR}	& 	\textbf{Hit@5}	&  \textbf{P@1} &	\textbf{MRR}	& 	\textbf{Hit@5}	 & \textbf{Robust}  \\ \toprule 
    \conquer~\cite{kaiser2021reinforcement}   & $0.218$ &	$0.272$	& $0.337$	 & $0.173$ &	$0.224$ &	$0.287$	& $6.531$ & $0.236$	& $0.287$ &	$0.360$ & $0.197$ &	$0.245$ &	$0.304$ & $6.447$	 \\
    \conquer~\cite{kaiser2021reinforcement} + \reign &  $\boldsymbol{0.245}$*	& $\boldsymbol{0.292}$*	& $\boldsymbol{0.346}$* &	$\boldsymbol{0.190}$* &	$\boldsymbol{0.236}$* &	$\boldsymbol{0.289}$*  & $\boldsymbol{7.035}$* & $\boldsymbol{0.238}$	& $\boldsymbol{0.290}$*	& $\boldsymbol{0.371}$*		 & $\boldsymbol{0.202}$*	& $\boldsymbol{0.252}$*	 & $\boldsymbol{0.310}$*	 & $\boldsymbol{7.224}$* \\ \midrule
    \explaignn~\cite{christmann2023explainable} & $0.370$ &	$0.438$ &	$0.526$ &	$0.278$	& $0.346$ &	$0.433$ & $10.983$ & $0.271$ &	$0.355$ &	$0.466$ &	$0.219$ &	$0.290$ &	$0.382$ & $8.400$\\
    \explaignn~\cite{christmann2023explainable} + \reign  & $\boldsymbol{0.384}$* &	$\boldsymbol{0.446}$* &	$\boldsymbol{0.531}$ &	$\boldsymbol{0.295}$* &	$\boldsymbol{0.361}$*  & $\boldsymbol{0.448}$*	& $\boldsymbol{11.130}$* &$\boldsymbol{0.318}$* & $\boldsymbol{0.411}$* &	$\boldsymbol{0.529}$*	&	$\boldsymbol{0.226}$* &	$\boldsymbol{0.302}$* & $\boldsymbol{0.402}$* & $\boldsymbol{8.925}$* \\
		 \bottomrule
	\end{tabular} }
	\caption{Main results comparing \reign-enhanced ConvQA models with their standalone versions. GPT-augmented test sets are 20x original sizes. \reign is applied zero-shot on \convquestions. The higher value per column per QA model is in bold.}
	\label{tab:main-res}
    \vspace*{-0.5cm}
\end{table*}

\myparagraph{\reign results in robust training} The four methods \conquer, \conquer + \reign (\conquer coupled with \reign), \explaignn, and \explaignn + \reign (\explaignn coupled with \reign) are evaluated on the two benchmarks \convmix and \convquestions. Results on test sets are in Table~\ref{tab:main-res}. A clear observation is that methods interfaced with \reign systematically outperform the original ConvQA models, on all test sets and metrics. While numbers are reported on the original test sets for completeness, results become much more significant on the GPT-test sets, with $p$-values of the order of $10^{-80}$ (recall that these values are averaged over $\simeq 100k$-$200k$ cases, Table~\ref{tab:data}). 
Importantly, versions with \reign score systematically higher on the \textit{robustness metric} (Sec.~\ref{sec:setup}), showing that the improved models are capable of handling more lexical and syntactic variations on average (differences higher for larger GPT-sets).
ConvQA with these benchmarks and GPT reformulations are challenging: these values are far less than $21$ (the \textit{Robust} measure here lies between $0$ and the number of reformulations per question including the original formulation, $21$). 
We also computed the number of unique \textit{intents} that \textit{newly become answerable} (P@1 $= 1$ for at least one question or one of its reformulations) with \reign: this is $115$ (\convmix-GPT-set) and $407$ (\convquestions-GPT-set) for \conquer, showing that our robust training can put more unique information needs within reach of the ConvQA model. 
Representative reformulations generated by \reign and GPT are in Tables~\ref{tab:bart-reformulations} and Table~\ref{tab:gpt-reformulations}, respectively. On average, original questions, \reign, and GPT-reformulations, are $5.9$, $7.5$, and $7.2$ words long.

\myparagraph{\reign components are generalizable} 
Results on the \convquestions benchmark showcase successful zero-shot application of \reign modules. Given that the \convquestions test sets are much larger than \convmix (see Table~\ref{tab:data}), improved results over the original QA modules show that our RCS and RG modules, individually, are immune to idiosyncrasies in specific datasets.

\myparagraph{Benefits of \reign hold over domains and turns} We report drill-down results over five domains and individual conversation turns in Tables~\ref{tab:domain-res} and~\ref{tab:turn-res}. We show that the benefits provided by reinforced reformulation generation are not limited to specific domains, or shallow conversation turns only.
\begin{table} [t] 
	\newcolumntype{G}{>{\columncolor [gray] {0.90}}c}
	\resizebox{\columnwidth}{!}{
	\begin{tabular}{l G  c G  c G} 
		\toprule
       \textbf{Method $\downarrow$ / Domain $\rightarrow$}   &     \textbf{Books} &\textbf{Movies} & \textbf{Music}    & \textbf{TV series}    & \textbf{Soccer} \\ \toprule 
    \conquer~\cite{kaiser2021reinforcement}   & $0.227$ & $0.175$ & $0.159$ & $0.141$ & $0.163$ \\
    \conquer~\cite{kaiser2021reinforcement} + \reign & $\boldsymbol{0.239}$* & $\boldsymbol{0.200}$* & $\boldsymbol{0.167}$* & $\boldsymbol{0.160}$* & $\boldsymbol{0.184}$*   \\ \midrule
    \explaignn~\cite{christmann2023explainable}  &	$0.298$ & $\boldsymbol{0.287}$* & $0.265$ & $0.274$ & $0.265$ 	 \\
    \explaignn~\cite{christmann2023explainable} + \reign &	$\boldsymbol{0.333}$* & $0.283$ & $\boldsymbol{0.301}$* & $\boldsymbol{0.281}$* & $\boldsymbol{0.275}$*    \\
		 \bottomrule
	\end{tabular} }
	\caption{Domain-wise P@1 results on GPT-\convmix testset.} 
	\label{tab:domain-res}
    \vspace{-0.7cm}
\end{table}

\begin{table} [t] 
	\newcolumntype{G}{>{\columncolor [gray] {0.90}}c}
	\resizebox{\columnwidth}{!}{
	\begin{tabular}{l G c G c  G c } 
		\toprule
       \textbf{Method $\downarrow$ / Turn $\rightarrow$} &   \textbf{1} &\textbf{2} & \textbf{3}    & \textbf{4}    &  \textbf{5} &\textbf{6-10}  \\ \toprule 
    \conquer~\cite{kaiser2021reinforcement}   & $0.205$ & $0.193$ & $0.177$ & $0.184$ & $0.160$ & $0.133$			   \\
    \conquer~\cite{kaiser2021reinforcement} + \reign & $\boldsymbol{0.210}$* & $\boldsymbol{0.214}$* & $\boldsymbol{0.194}$* & $\boldsymbol{0.204}$* & $\boldsymbol{0.184}$* & $\boldsymbol{0.147}$*	    \\ \midrule
    \explaignn~\cite{christmann2023explainable}  & $0.333$ & $0.297$ & $0.286$ & $0.292$ & $0.277$ & $0.205$			 	 \\
    \explaignn~\cite{christmann2023explainable} + \reign & $\boldsymbol{0.350}$* & $\boldsymbol{0.318}$* & $\boldsymbol{0.311}$* & $\boldsymbol{0.305}$* & $\boldsymbol{0.291}$* & $\boldsymbol{0.216}$*   \\
		 \bottomrule
	\end{tabular} }
	\caption{Turn-wise P@1 results on GPT-\convmix testset.} 
	\label{tab:turn-res}
    \vspace*{-0.7cm}
\end{table}

\subsection{In-depth analysis}
\label{subsec:indepth}

\begin{table} [t] 
	\newcolumntype{G}{>{\columncolor [gray] {0.90}}c}
    \newcolumntype{H}{>{\setbox0=\hbox\bgroup}c<{\egroup}@{}}
	\resizebox{\columnwidth}{!}{
	\begin{tabular}{c l G G G c} 
    \toprule
    	\textbf{Row}   & \textbf{Configuration}	   &	\textbf{P@1}  &	\textbf{MRR}	& 	\textbf{Hit@5}	& \textbf{\#Data} \\\toprule 
        $1$ & RCS (DQN, top-5) + RG (BART) [Full]  &  $\boldsymbol{0.190}$ &	$\boldsymbol{0.236}$ &	$0.289$  & $43.6$k \\ \midrule
        $2$ & RCS (DQN, top-3) + RG (BART)  &   $0.184$ &	$0.231$ &$0.288$  & $30.5$k \\ 
        $3$ & RCS (DQN, top-1) + RG (BART)  &	 $0.178$ &	$0.228$ &	$0.288$  & $15.9$k \\ \midrule 
        $4$ & No RCS (All cats) +  RG (BART)       &  	$0.188$ & 	$0.234$ &	$0.292$ & $126$k    \\
        $5$ & No RCS (Random cats) +  RG (BART)  &   $0.182$ &	$0.232$ &	$\boldsymbol{0.293}$ & $42$k\\
        $6$ & No RCS (Sample cats) +  RG (BART)  &  	$0.185$ &	$0.231$	& $0.287$ & $41.9$k \\ \midrule
        $7$ & No RCS (INS part) + RG (BART)   &  $0.183$ &	$0.230$ &	$0.288$ & $42$k\\
        $8$ & No RCS (DEL part) + RG (BART) &   $0.172$ &	$0.218$ &	$0.273$ & $42$k\\
        $9$ & No RCS (SUBS part) +  RG (BART)  &   $0.183$ &	$0.228$ &	$0.282$ & $58.8$k\\ \midrule
        $10$ & No RCS + No RG (Question completion) &   $0.175$ &	$0.224$ &	$0.284$ & $15.1$k\\
        $11$ & No RCS + No RG (Question rewriting) &   $0.180$ &	$0.230$ &	$0.291$ & $15.1$k\\ \midrule
	\end{tabular} }
	\caption{Large-scale effects of design choices in \reign (with \conquer on GPT-\convmix, all differences systematic).}
	\label{tab:ablation}
    \vspace*{-0.7cm}
\end{table}

In Table~\ref{tab:ablation}, we report in-depth analyses of the moving parts in \reign, using \conquer on the \convmix-GPT-set. Trends with \explaignn and \convquestions are similar. We do not use this table for making design choices -- rather, we expose large-scale effects of sub-optimal configurations: hence the choice of a $\simeq 100$k-GPT set instead of the typically small dev set.

\myparagraph{RCS with DQN is vital} First and foremost, we show that selecting reformulations with our DQN is necessary, and simply taking \textit{all} noisy reformulations does not serve as a sledgehammer for performance improvement even at three times the number of data points used (Row 1 vs. Row 4).
This makes a solid case for judicious augmentation. Using all reformulations does lead to higher answer recall as seen through the Hit@5 value, but at the cost of precise ranking.
Using top-5 reformulations is a sweet spot for deploying the RCS (Row 1 vs. 2 and 3). Using higher numbers drastically increases the training time and often produces degenerate reformulations. Contrast against a \textit{random} choice of categories inside the RCS is a natural experiment, and we find this to be sub-optimal (see P@1 in Row 5). Another stronger baseline is to \textit{sample} $k=5$ categories according to the Q-value distribution: this again falls short of a top-$k$ prediction (Row 6).

\begin{figure} [t]
	\centering
	\includegraphics[width=\columnwidth]{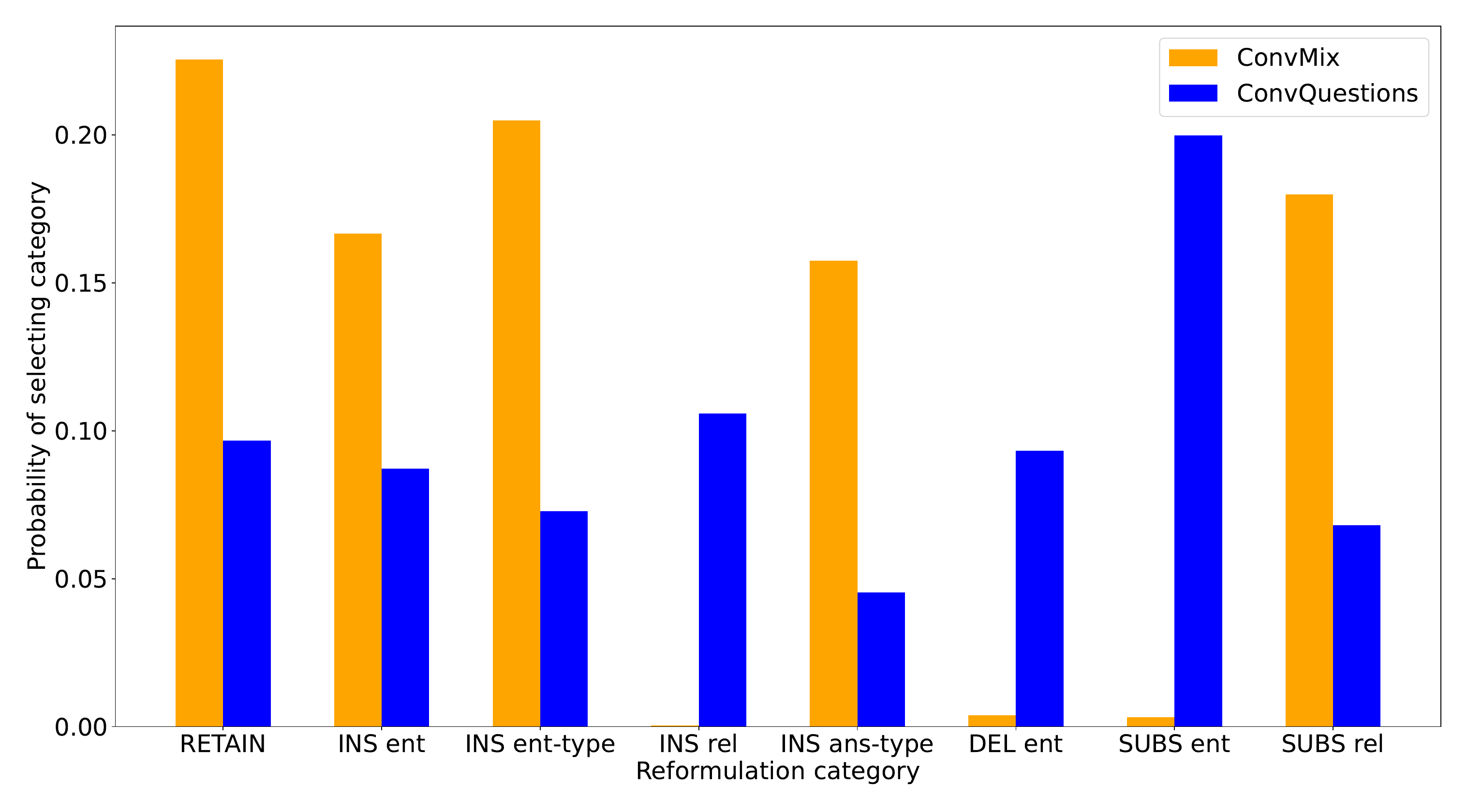}
    \vspace*{-0.7cm}
	\caption{Common category predictions by the RCS DQN.} 
	\label{fig:categories}
    \vspace*{-0.3cm}
\end{figure}

\myparagraph{The whole taxonomy matters} It may appear that using only insertion or substitution operations from the taxonomy may suffice for robust learning, but we find that considering all categories jointly (Row 1) is superior to using only individual ``meta''-categories (INS, DEL SUBS in Rows $7-9$). While using \textit{only} deletion operations hurts performance the most (Row 8), it is thus clear that carefully removing parts of questions also contributes to a stronger model (for example, deleting an entity was considered to improve MRR $10\%$ of the time on \convquestions, presumably removing noise). Fig.~\ref{fig:categories} shows the union of the top-5 frequent predictions from our RCS DQN for the two benchmarks. Insertion of question entity types and expected answer types are generally useful for disambiguation, and substituting relations with aliases naturally makes the system more robust to predicate paraphrasing. The original question was retained $10-20\%$ of the time.

\myparagraph{Question rewriting is not enough} As discussed in Sec.~\ref{sec:intro}, reformulating a conversational question into a more complete form at \textit{answering time} is a prevalent approach in ConvQA. As such, comparison with such rewriting or completion approaches is out of scope, as we focus on more robust \textit{training}. Nevertheless, we explore the natural possibility of using completed forms of questions during training, as opposed to a set of noisy reformulations. The \convmix benchmark~\cite{christmann2022conversational} contains intent-explicit questions by the original crowdworkers who generated the conversations, and can thus be treated as gold standard completions. We found that this falls short of our proposed method (Row 1 vs. Row 10), as does question rewriting using T5~\cite{lin2021multi} (Row 1 vs. Row 11). Interestingly, corroborating findings from the BART experiments, noisy rewrites with T5 outperform human completions. Note that completion or rewriting entails \textit{one} longer version of the question (hence $\simeq 15$k data points): we find that generating a \textit{small set} of potentially incomplete variants with \reign improves performance.

\myparagraph{Intrinsic rewards also work well} Our DQN uses differences in reciprocal ranks, computed from gold answers in benchmarks, as \textit{extrinsic} rewards. A natural question is what happens in cases where such relevance assessments are not available. We thus explored an alternative of an \textit{intrinsic} reward~\cite{liu2022rainier,chen2022reinforced} computed as the \textit{difference} in the ConvQA \textit{model's probabilities} of its top-1 answer for the reformulation and the original question~\cite{liu2022rainier,chen2022reinforced} (this is analogous to blind relevance feedback). On a positive note, this resulted in comparable performance on the \convmix dev set ($0.270$ P@1 for extrinsic vs. $0.269$ for intrinsic; $0.311$ MRR for both).

\myparagraph{Manual error analysis} The authors analyzed $10$ reformulations from each category for both BART reformulations and the original fine-tuning data ($15 \times 10 \times 2 = 300$ in all), to look for potential issues. There were only minor problems detected for both scenarios.
The concerns with BART were as follows: unintelligible intent (4 cases), hallucinations (5), wrong category applied (13), information removed unintentionally (15), transformation possible but not made (13), unsuitable entity or type added (4), and information already in the question was added again (5). The concerns with the initial noisy data can sometimes be traced back to incorrect processing of the benchmarks (Sec.~\ref{subsec:rg-train}), like wrong predicate (4 cases) and addition of information already present due to incorrect markup (4).
Other errors include: intent changed (6), unsuitable types inserted (7 cases), and changes possible but not made (2).

\myparagraph{GPT cannot replace the \reign pipeline} It is a common trend nowadays to use LLMs like GPT at multiple points in pipelines. We thus checked whether the same ChatGPT model that generated our test set could actually replace the whole \reign pipeline by directly generating reformulations for training questions. Importantly, we found that this underperforms \reign when five reformulations are considered for each alternative, on the original \convmix dev set (evaluation of GPT reformulations on GPT test sets could result in undesirable biases): $0.270$ P@1 for \reign vs. $0.261$ for GPT (\conquer), and $0.423$ for \reign vs. $0.405$ for GPT (\explaignn). Note that the GPT reformulations are \textit{model-agnostic}: this shows that reformulations generated with \textit{model-aware performance feedback} is indeed a better choice for robust training.

%% file: sections/06-related.tex
\section{Related Work}
\label{sec:related}

\myparagraph{Conversational question answering} ConvQA~\cite{choi2018quac,reddy2019coqa,saha2018complex,christmann2019look,saharoy2022question} can be viewed as a research direction under the umbrella of conversational search~\cite{zamani2022conversational,dalton2022conversational,owoicho2023exploiting}, with natural-language utterances as input. Answers are crisp entities~\cite{guo2018dialog,kacupaj2022contrastive}, sentences~\cite{baheti2020fluent}, or passages~\cite{dalton2019cast,qu2020open}. 
Methods proposed belong to two major families; they either (i) derive a \textit{self-contained version} of the question that can be handled by a standard QA system (referred to as rewriting~\cite{vakulenko2021question,yu2020few,ke2022knowledge,chen2022reinforced,ishii2022integrating,raposo2022question}, resolution~\cite{kim2021learn,voskarides2020query}, or even reformulation~\cite{vakulenko2020wrong,mo2023convgqr}, in different works), or (ii) 
\textit{model the history} as additional context to answer the current question~\cite{qu2019bert,gekhman2023robustness,qu2019attentive,gupta2021role,qiu2021reinforced,kaiser2021reinforcement,sun2023history}. \reign is not a QA model by itself, but can improve the performance of any given ConvQA system: we demonstrate this by choosing one method from each family of approaches in our experiments~\cite{kaiser2021reinforcement,christmann2023explainable}.
In this work, we enhance conversational QA over KGs~\cite{guo2018dialog,shen2019multi,kacupaj2022contrastive,kacupaj2021conversational,perez2023semantic,jain2023conversational}, where 
answers are small sets of entities.

\myparagraph{Robustness in QA} Improving the robustness or generalizability of ConvQA models has not seen much dedicated activity: work has mostly been limited to specific benchmarks of choice~\cite{saha2018complex,guo2018dialog,shen2019multi,christmann2019look,christmann2022conversational}. Implicitly, authors have tried to prove robust behavior by the use of multiple benchmarks~\cite{kacupaj2022contrastive,marion2021structured,lan2021modeling}, or zero-shot application of models to new benchmarks~\cite{christmann2023explainable}. \textit{Data augmentation}, given one or more benchmarks, is one of the prominent approaches for increasing model robustness in QA~\cite{bartolo2021improving,puri2020training,liu2020tell,sachan2018self,baheti2020fluent,shen2022product,yang2019data}. Our work stands out as model-specific data augmentation, a philosophy for effective training by trying to fill ``gaps'' in a specific model's learned behavior, instead of feeding in a very large volume of noisy data to all models. Some recent works in QA over text investigate model robustness by perturbing input passages~\cite{gekhman2023robustness,neeman2022disentqa}, while we tap into question reformulations as a perturbation on the question-side.

\myparagraph{Reformulations in search and QA} Work on question or query reformulations in search~\cite{li2022cooperative,xue2012automatically,ponnusamy2020feedback,nogueira2017task}, QA~\cite{tomuro2003interrogative,liu2008query,hermjakob2002natural,buck2018ask,kaiser2021reinforcement}, and recommenders~\cite{zhang2022analyzing}, can be broadly positioned in a $2 \times 2$ space according to the definition of a reformulation:\squishlist
\item Rephrasing (of the same intent) \cite{kaiser2021reinforcement,dong2017learning,fader2013paraphrase,tomuro2001selecting} \textit{versus} refinement (into a variation of the previous intent) \cite{liu2008query,zhang2022analyzing,nogueira2017task,liu2019generative}.
\item Using for better training \cite{li2022cooperative,ponnusamy2020feedback} \textit{versus} using for better inference~\cite{buck2018ask,xue2012automatically,nogueira2017task,das2019multi}. 
\squishend
This work falls into the \textit{rephrasing-for-training} quadrant, viewing reformulations as rephrased user utterances for the current question in a conversation,
and leveraging these for training a more robust model. Early work on automatic acquisition of query reformulation patterns~\cite{xue2012automatically,tomuro2003interrogative,liu2008query,tomuro2001selecting},
or on \textit{paraphrasing} for improving model robustness~\cite{abujabal2018never,fader2014open,fader2013paraphrase,dong2017learning,berant2014semantic,abujabal19comqa,gan2019improving}, did not account for answers from previous turns, and more generally, did not address the specific difficulty of incomplete and ad-hoc user utterances in conversations.

%% file: sections/07-conclusion.tex
\section{Conclusion}
\label{sec:confut}

This work contributes a method that makes conversational question answering models more robust with generated reformulations that are specifically \textit{guided} towards better QA performance. The proposed framework
judiciously picks the most suitable choices for enhanced training, as opposed to brute-force data augmentation with all possible reformulations. Experiments with two state-of-the-art ConvQA methods
demonstrate benefits of the \reign method.

\section*{Acknowledgments}
\label{sec:ack}

We thank Philipp Christmann from the Max Planck Institute for Informatics for useful inputs at various stages of this work.

%% file: sections/08-ethics.tex
\section*{Ethical considerations}
\label{sec:ethics}

There are no negative ethical and societal concerns arising from this work.
The questions and reformulations are based on public benchmarks, and do not contain any sensitive or personally identifiable information (PII).
Prompts for ChatGPT are not meant to evoke adversarial, hateful, or malicious responses. Security, safety, and fairness concerns are not applicable as well.

%% file: 2024-wsdm-fp-reign.bbl

\begin{thebibliography}{87}


\ifx \showCODEN    \undefined \def \showCODEN     #1{\unskip}     \fi
\ifx \showDOI      \undefined \def \showDOI       #1{#1}\fi
\ifx \showISBNx    \undefined \def \showISBNx     #1{\unskip}     \fi
\ifx \showISBNxiii \undefined \def \showISBNxiii  #1{\unskip}     \fi
\ifx \showISSN     \undefined \def \showISSN      #1{\unskip}     \fi
\ifx \showLCCN     \undefined \def \showLCCN      #1{\unskip}     \fi
\ifx \shownote     \undefined \def \shownote      #1{#1}          \fi
\ifx \showarticletitle \undefined \def \showarticletitle #1{#1}   \fi
\ifx \showURL      \undefined \def \showURL       {\relax}        \fi
\providecommand\bibfield[2]{#2}
\providecommand\bibinfo[2]{#2}
\providecommand\natexlab[1]{#1}
\providecommand\showeprint[2][]{arXiv:#2}

\bibitem[Abujabal et~al\mbox{.}(2018)]%
        {abujabal2018never}
\bibfield{author}{\bibinfo{person}{Abdalghani Abujabal},
  \bibinfo{person}{Rishiraj Saha~Roy}, \bibinfo{person}{Mohamed Yahya}, {and}
  \bibinfo{person}{Gerhard Weikum}.} \bibinfo{year}{2018}\natexlab{}.
\newblock \showarticletitle{Never-ending learning for open-domain question
  answering over knowledge bases}. In \bibinfo{booktitle}{\emph{WWW}}.
\newblock


\bibitem[Abujabal et~al\mbox{.}(2019)]%
        {abujabal19comqa}
\bibfield{author}{\bibinfo{person}{Abdalghani Abujabal},
  \bibinfo{person}{Rishiraj Saha~Roy}, \bibinfo{person}{Mohamed Yahya}, {and}
  \bibinfo{person}{Gerhard Weikum}.} \bibinfo{year}{2019}\natexlab{}.
\newblock \showarticletitle{{ComQA: A Community-sourced Dataset for Complex
  Factoid Question Answering with Paraphrase Clusters}}. In
  \bibinfo{booktitle}{\emph{NAACL-HLT '19}}.
\newblock


\bibitem[Anantha et~al\mbox{.}(2020)]%
        {anantha2020open}
\bibfield{author}{\bibinfo{person}{Raviteja Anantha}, \bibinfo{person}{Svitlana
  Vakulenko}, \bibinfo{person}{Zhucheng Tu}, \bibinfo{person}{Shayne Longpre},
  \bibinfo{person}{Stephen Pulman}, {and} \bibinfo{person}{Srinivas Chappidi}.}
  \bibinfo{year}{2020}\natexlab{}.
\newblock \showarticletitle{Open-Domain Question Answering Goes Conversational
  via Question Rewriting}. In \bibinfo{booktitle}{\emph{arXiv}}.
\newblock


\bibitem[Auer et~al\mbox{.}(2007)]%
        {auer2007dbpedia}
\bibfield{author}{\bibinfo{person}{S{\"o}ren Auer}, \bibinfo{person}{Christian
  Bizer}, \bibinfo{person}{Georgi Kobilarov}, \bibinfo{person}{Jens Lehmann},
  \bibinfo{person}{Richard Cyganiak}, {and} \bibinfo{person}{Zachary Ives}.}
  \bibinfo{year}{2007}\natexlab{}.
\newblock \showarticletitle{{DBpedia: A nucleus for a Web of open data}}.
\newblock \bibinfo{journal}{\emph{The Semantic Web}} (\bibinfo{year}{2007}).
\newblock


\bibitem[Baheti et~al\mbox{.}(2020)]%
        {baheti2020fluent}
\bibfield{author}{\bibinfo{person}{Ashutosh Baheti}, \bibinfo{person}{Alan
  Ritter}, {and} \bibinfo{person}{Kevin Small}.}
  \bibinfo{year}{2020}\natexlab{}.
\newblock \showarticletitle{Fluent Response Generation for Conversational
  Question Answering}. In \bibinfo{booktitle}{\emph{ACL}}.
\newblock


\bibitem[Bartolo et~al\mbox{.}(2021)]%
        {bartolo2021improving}
\bibfield{author}{\bibinfo{person}{Max Bartolo}, \bibinfo{person}{Tristan
  Thrush}, \bibinfo{person}{Robin Jia}, \bibinfo{person}{Sebastian Riedel},
  \bibinfo{person}{Pontus Stenetorp}, {and} \bibinfo{person}{Douwe Kiela}.}
  \bibinfo{year}{2021}\natexlab{}.
\newblock \showarticletitle{Improving Question Answering Model Robustness with
  Synthetic Adversarial Data Generation}. In \bibinfo{booktitle}{\emph{EMNLP}}.
\newblock


\bibitem[Bast and Haussmann(2015)]%
        {bast2015more}
\bibfield{author}{\bibinfo{person}{Hannah Bast} {and} \bibinfo{person}{Elmar
  Haussmann}.} \bibinfo{year}{2015}\natexlab{}.
\newblock \showarticletitle{{More accurate question answering on Freebase}}. In
  \bibinfo{booktitle}{\emph{CIKM}}.
\newblock


\bibitem[Berant and Liang(2014)]%
        {berant2014semantic}
\bibfield{author}{\bibinfo{person}{Jonathan Berant} {and}
  \bibinfo{person}{Percy Liang}.} \bibinfo{year}{2014}\natexlab{}.
\newblock \showarticletitle{Semantic parsing via paraphrasing}. In
  \bibinfo{booktitle}{\emph{ACL}}.
\newblock


\bibitem[Buck et~al\mbox{.}(2018)]%
        {buck2018ask}
\bibfield{author}{\bibinfo{person}{Christian Buck}, \bibinfo{person}{Jannis
  Bulian}, \bibinfo{person}{Massimiliano Ciaramita}, \bibinfo{person}{Wojciech
  Gajewski}, \bibinfo{person}{Andrea Gesmundo}, \bibinfo{person}{Neil Houlsby},
  {and} \bibinfo{person}{Wei Wang}.} \bibinfo{year}{2018}\natexlab{}.
\newblock \showarticletitle{Ask the right questions: {A}ctive question
  reformulation with reinforcement learning}. In
  \bibinfo{booktitle}{\emph{ICLR}}.
\newblock


\bibitem[Chen et~al\mbox{.}(2022)]%
        {chen2022reinforced}
\bibfield{author}{\bibinfo{person}{Zhiyu Chen}, \bibinfo{person}{Jie Zhao},
  \bibinfo{person}{Anjie Fang}, \bibinfo{person}{Besnik Fetahu},
  \bibinfo{person}{Oleg Rokhlenko}, {and} \bibinfo{person}{Shervin Malmasi}.}
  \bibinfo{year}{2022}\natexlab{}.
\newblock \showarticletitle{Reinforced question rewriting for conversational
  question answering}.
\newblock \bibinfo{journal}{\emph{arXiv}} (\bibinfo{year}{2022}).
\newblock


\bibitem[Choi et~al\mbox{.}(2018)]%
        {choi2018quac}
\bibfield{author}{\bibinfo{person}{Eunsol Choi}, \bibinfo{person}{He He},
  \bibinfo{person}{Mohit Iyyer}, \bibinfo{person}{Mark Yatskar},
  \bibinfo{person}{Wen-tau Yih}, \bibinfo{person}{Yejin Choi},
  \bibinfo{person}{Percy Liang}, {and} \bibinfo{person}{Luke Zettlemoyer}.}
  \bibinfo{year}{2018}\natexlab{}.
\newblock \showarticletitle{{QuAC: Q}uestion answering in context}. In
  \bibinfo{booktitle}{\emph{EMNLP}}.
\newblock


\bibitem[Christmann et~al\mbox{.}(2019)]%
        {christmann2019look}
\bibfield{author}{\bibinfo{person}{Philipp Christmann},
  \bibinfo{person}{Rishiraj Saha~Roy}, \bibinfo{person}{Abdalghani Abujabal},
  \bibinfo{person}{Jyotsna Singh}, {and} \bibinfo{person}{Gerhard Weikum}.}
  \bibinfo{year}{2019}\natexlab{}.
\newblock \showarticletitle{Look before you Hop: Conversational Question
  Answering over Knowledge Graphs Using Judicious Context Expansion}. In
  \bibinfo{booktitle}{\emph{CIKM}}.
\newblock


\bibitem[Christmann et~al\mbox{.}(2022a)]%
        {christmann2022beyond}
\bibfield{author}{\bibinfo{person}{Philipp Christmann},
  \bibinfo{person}{Rishiraj Saha~Roy}, {and} \bibinfo{person}{Gerhard Weikum}.}
  \bibinfo{year}{2022}\natexlab{a}.
\newblock \showarticletitle{Beyond NED: Fast and Effective Search Space
  Reduction for Complex Question Answering over Knowledge Bases}. In
  \bibinfo{booktitle}{\emph{WSDM}}.
\newblock


\bibitem[Christmann et~al\mbox{.}(2022b)]%
        {christmann2022conversational}
\bibfield{author}{\bibinfo{person}{Philipp Christmann},
  \bibinfo{person}{Rishiraj Saha~Roy}, {and} \bibinfo{person}{Gerhard Weikum}.}
  \bibinfo{year}{2022}\natexlab{b}.
\newblock \showarticletitle{Conversational Question Answering on Heterogeneous
  Sources}. In \bibinfo{booktitle}{\emph{SIGIR}}.
\newblock


\bibitem[Christmann et~al\mbox{.}(2023)]%
        {christmann2023explainable}
\bibfield{author}{\bibinfo{person}{Philipp Christmann},
  \bibinfo{person}{Rishiraj Saha~Roy}, {and} \bibinfo{person}{Gerhard Weikum}.}
  \bibinfo{year}{2023}\natexlab{}.
\newblock \showarticletitle{Explainable Conversational Question Answering over
  Heterogeneous Sources via Iterative Graph Neural Networks}. In
  \bibinfo{booktitle}{\emph{SIGIR}}.
\newblock


\bibitem[Dalton et~al\mbox{.}(2022)]%
        {dalton2022conversational}
\bibfield{author}{\bibinfo{person}{Jeffrey Dalton}, \bibinfo{person}{Sophie
  Fischer}, \bibinfo{person}{Paul Owoicho}, \bibinfo{person}{Filip Radlinski},
  \bibinfo{person}{Federico Rossetto}, \bibinfo{person}{Johanne~R Trippas},
  {and} \bibinfo{person}{Hamed Zamani}.} \bibinfo{year}{2022}\natexlab{}.
\newblock \showarticletitle{Conversational Information Seeking: Theory and
  Application}. In \bibinfo{booktitle}{\emph{SIGIR}}.
\newblock


\bibitem[Dalton et~al\mbox{.}(2019)]%
        {dalton2019cast}
\bibfield{author}{\bibinfo{person}{Jeffrey Dalton}, \bibinfo{person}{Chenyan
  Xiong}, {and} \bibinfo{person}{Jamie Callan}.}
  \bibinfo{year}{2019}\natexlab{}.
\newblock \showarticletitle{{CAsT 2019: The conversational assistance track
  overview}}. In \bibinfo{booktitle}{\emph{TREC}}.
\newblock


\bibitem[Das et~al\mbox{.}(2019)]%
        {das2019multi}
\bibfield{author}{\bibinfo{person}{Rajarshi Das}, \bibinfo{person}{Shehzaad
  Dhuliawala}, \bibinfo{person}{Manzil Zaheer}, {and} \bibinfo{person}{Andrew
  McCallum}.} \bibinfo{year}{2019}\natexlab{}.
\newblock \showarticletitle{Multi-step retriever-reader interaction for
  scalable open-domain question answering}. In
  \bibinfo{booktitle}{\emph{ICLR}}.
\newblock


\bibitem[Devlin et~al\mbox{.}(2019)]%
        {devlin2018bert}
\bibfield{author}{\bibinfo{person}{Jacob Devlin}, \bibinfo{person}{Ming-Wei
  Chang}, \bibinfo{person}{Kenton Lee}, {and} \bibinfo{person}{Kristina
  Toutanova}.} \bibinfo{year}{2019}\natexlab{}.
\newblock \showarticletitle{{BERT: Pre-training of deep bidirectional
  transformers for language understanding}}. In
  \bibinfo{booktitle}{\emph{NAACL-HLT}}.
\newblock


\bibitem[Dong et~al\mbox{.}(2017)]%
        {dong2017learning}
\bibfield{author}{\bibinfo{person}{Li Dong}, \bibinfo{person}{Jonathan
  Mallinson}, \bibinfo{person}{Siva Reddy}, {and} \bibinfo{person}{Mirella
  Lapata}.} \bibinfo{year}{2017}\natexlab{}.
\newblock \showarticletitle{Learning to Paraphrase for Question Answering}. In
  \bibinfo{booktitle}{\emph{EMNLP}}.
\newblock


\bibitem[Fader et~al\mbox{.}(2013)]%
        {fader2013paraphrase}
\bibfield{author}{\bibinfo{person}{Anthony Fader}, \bibinfo{person}{Luke
  Zettlemoyer}, {and} \bibinfo{person}{Oren Etzioni}.}
  \bibinfo{year}{2013}\natexlab{}.
\newblock \showarticletitle{Paraphrase-driven learning for open question
  answering}. In \bibinfo{booktitle}{\emph{ACL}}.
\newblock


\bibitem[Fader et~al\mbox{.}(2014)]%
        {fader2014open}
\bibfield{author}{\bibinfo{person}{Anthony Fader}, \bibinfo{person}{Luke
  Zettlemoyer}, {and} \bibinfo{person}{Oren Etzioni}.}
  \bibinfo{year}{2014}\natexlab{}.
\newblock \showarticletitle{Open question answering over curated and extracted
  knowledge bases}. In \bibinfo{booktitle}{\emph{KDD}}.
\newblock


\bibitem[Ferragina and Scaiella(2010)]%
        {ferragina2010tagme}
\bibfield{author}{\bibinfo{person}{Paolo Ferragina} {and} \bibinfo{person}{Ugo
  Scaiella}.} \bibinfo{year}{2010}\natexlab{}.
\newblock \showarticletitle{{TAGME: On-the-fly annotation of short text
  fragments (by Wikipedia entities)}}. In \bibinfo{booktitle}{\emph{CIKM}}.
  \bibinfo{pages}{1625--1628}.
\newblock


\bibitem[Gan and Ng(2019)]%
        {gan2019improving}
\bibfield{author}{\bibinfo{person}{Wee~Chung Gan} {and}
  \bibinfo{person}{Hwee~Tou Ng}.} \bibinfo{year}{2019}\natexlab{}.
\newblock \showarticletitle{Improving the robustness of question answering
  systems to question paraphrasing}. In \bibinfo{booktitle}{\emph{ACL}}.
\newblock


\bibitem[Gekhman et~al\mbox{.}(2023)]%
        {gekhman2023robustness}
\bibfield{author}{\bibinfo{person}{Zorik Gekhman}, \bibinfo{person}{Nadav
  Oved}, \bibinfo{person}{Orgad Keller}, \bibinfo{person}{Idan Szpektor}, {and}
  \bibinfo{person}{Roi Reichart}.} \bibinfo{year}{2023}\natexlab{}.
\newblock \showarticletitle{On the Robustness of Dialogue History
  Representation in Conversational Question Answering: A Comprehensive Study
  and a New Prompt-based Method}.
\newblock  (\bibinfo{year}{2023}).
\newblock


\bibitem[Guo et~al\mbox{.}(2018)]%
        {guo2018dialog}
\bibfield{author}{\bibinfo{person}{Daya Guo}, \bibinfo{person}{Duyu Tang},
  \bibinfo{person}{Nan Duan}, \bibinfo{person}{Ming Zhou}, {and}
  \bibinfo{person}{Jian Yin}.} \bibinfo{year}{2018}\natexlab{}.
\newblock \showarticletitle{{Dialog-to-action: C}onversational question
  answering over a large-scale knowledge base}. In
  \bibinfo{booktitle}{\emph{NeurIPS}}.
\newblock


\bibitem[Gupta and Sharma(2021)]%
        {gupta2021role}
\bibfield{author}{\bibinfo{person}{Somil Gupta} {and} \bibinfo{person}{Neeraj
  Sharma}.} \bibinfo{year}{2021}\natexlab{}.
\newblock \showarticletitle{Role of Attentive History Selection in
  Conversational Information Seeking}. In \bibinfo{booktitle}{\emph{arXiv}}.
\newblock


\bibitem[Hermjakob et~al\mbox{.}(2002)]%
        {hermjakob2002natural}
\bibfield{author}{\bibinfo{person}{Ulf Hermjakob}, \bibinfo{person}{Abdessamad
  Echihabi}, {and} \bibinfo{person}{Daniel Marcu}.}
  \bibinfo{year}{2002}\natexlab{}.
\newblock \showarticletitle{Natural Language Based Reformulation Resource and
  Wide Exploitation for Question Answering.}. In
  \bibinfo{booktitle}{\emph{TREC}}.
\newblock


\bibitem[Hirschman and Gaizauskas(2001)]%
        {hirschman2001natural}
\bibfield{author}{\bibinfo{person}{Lynette Hirschman} {and}
  \bibinfo{person}{Robert Gaizauskas}.} \bibinfo{year}{2001}\natexlab{}.
\newblock \showarticletitle{{Natural language question answering: The view from
  here}}.
\newblock \bibinfo{journal}{\emph{Natural Language Engineering}}
  \bibinfo{volume}{7}, \bibinfo{number}{4} (\bibinfo{year}{2001}),
  \bibinfo{pages}{275--300}.
\newblock


\bibitem[Ishii et~al\mbox{.}(2022a)]%
        {ishii2022integrating}
\bibfield{author}{\bibinfo{person}{Etsuko Ishii}, \bibinfo{person}{Bryan
  Wilie}, \bibinfo{person}{Yan Xu}, \bibinfo{person}{Samuel Cahyawijaya}, {and}
  \bibinfo{person}{Pascale Fung}.} \bibinfo{year}{2022}\natexlab{a}.
\newblock \showarticletitle{Integrating Question Rewrites in Conversational
  Question Answering: A Reinforcement Learning Approach}. In
  \bibinfo{booktitle}{\emph{ACL Student Research Workshop}}.
\newblock


\bibitem[Ishii et~al\mbox{.}(2022b)]%
        {ishii2022can}
\bibfield{author}{\bibinfo{person}{Etsuko Ishii}, \bibinfo{person}{Yan Xu},
  \bibinfo{person}{Samuel Cahyawijaya}, {and} \bibinfo{person}{Bryan Wilie}.}
  \bibinfo{year}{2022}\natexlab{b}.
\newblock \showarticletitle{Can Question Rewriting Help Conversational Question
  Answering?}. In \bibinfo{booktitle}{\emph{3rd Workshop on Insights from
  Negative Results in NLP}}.
\newblock


\bibitem[Jain and Lapata(2023)]%
        {jain2023conversational}
\bibfield{author}{\bibinfo{person}{Parag Jain} {and} \bibinfo{person}{Mirella
  Lapata}.} \bibinfo{year}{2023}\natexlab{}.
\newblock \showarticletitle{Conversational Semantic Parsing using Dynamic
  Context Graphs}.
\newblock \bibinfo{journal}{\emph{arXiv preprint arXiv:2305.06164}}
  (\bibinfo{year}{2023}).
\newblock


\bibitem[Kacupaj et~al\mbox{.}(2021)]%
        {kacupaj2021conversational}
\bibfield{author}{\bibinfo{person}{Endri Kacupaj}, \bibinfo{person}{Joan
  Plepi}, \bibinfo{person}{Kuldeep Singh}, \bibinfo{person}{Harsh Thakkar},
  \bibinfo{person}{Jens Lehmann}, {and} \bibinfo{person}{Maria Maleshkova}.}
  \bibinfo{year}{2021}\natexlab{}.
\newblock \showarticletitle{Conversational Question Answering over Knowledge
  Graphs with Transformer and Graph Attention Networks}. In
  \bibinfo{booktitle}{\emph{EACL}}.
\newblock


\bibitem[Kacupaj et~al\mbox{.}(2022)]%
        {kacupaj2022contrastive}
\bibfield{author}{\bibinfo{person}{Endri Kacupaj}, \bibinfo{person}{Kuldeep
  Singh}, \bibinfo{person}{Maria Maleshkova}, {and} \bibinfo{person}{Jens
  Lehmann}.} \bibinfo{year}{2022}\natexlab{}.
\newblock \showarticletitle{Contrastive Representation Learning for
  Conversational Question Answering over Knowledge Graphs}. In
  \bibinfo{booktitle}{\emph{CIKM}}.
\newblock


\bibitem[Kaiser et~al\mbox{.}(2020)]%
        {kaiser2020conversational}
\bibfield{author}{\bibinfo{person}{Magdalena Kaiser}, \bibinfo{person}{Rishiraj
  Saha~Roy}, {and} \bibinfo{person}{Gerhard Weikum}.}
  \bibinfo{year}{2020}\natexlab{}.
\newblock \showarticletitle{Conversational Question Answering over Passages by
  Leveraging Word Proximity Networks}. In \bibinfo{booktitle}{\emph{SIGIR}}.
\newblock


\bibitem[Kaiser et~al\mbox{.}(2021)]%
        {kaiser2021reinforcement}
\bibfield{author}{\bibinfo{person}{Magdalena Kaiser}, \bibinfo{person}{Rishiraj
  Saha~Roy}, {and} \bibinfo{person}{Gerhard Weikum}.}
  \bibinfo{year}{2021}\natexlab{}.
\newblock \showarticletitle{Reinforcement learning from reformulations in
  conversational question answering over knowledge graphs}. In
  \bibinfo{booktitle}{\emph{SIGIR}}.
\newblock


\bibitem[Ke et~al\mbox{.}(2022)]%
        {ke2022knowledge}
\bibfield{author}{\bibinfo{person}{Xirui Ke}, \bibinfo{person}{Jing Zhang},
  \bibinfo{person}{Xin Lv}, \bibinfo{person}{Yiqi Xu}, \bibinfo{person}{Shulin
  Cao}, \bibinfo{person}{Cuiping Li}, \bibinfo{person}{Hong Chen}, {and}
  \bibinfo{person}{Juanzi Li}.} \bibinfo{year}{2022}\natexlab{}.
\newblock \showarticletitle{Knowledge-augmented Self-training of A Question
  Rewriter for Conversational Knowledge Base Question Answering}. In
  \bibinfo{booktitle}{\emph{EMNLP}}.
\newblock


\bibitem[Kim et~al\mbox{.}(2021)]%
        {kim2021learn}
\bibfield{author}{\bibinfo{person}{Gangwoo Kim}, \bibinfo{person}{Hyunjae Kim},
  \bibinfo{person}{Jungsoo Park}, {and} \bibinfo{person}{Jaewoo Kang}.}
  \bibinfo{year}{2021}\natexlab{}.
\newblock \showarticletitle{{Learn to Resolve Conversational Dependency: A
  Consistency Training Framework for Conversational Question Answering}}. In
  \bibinfo{booktitle}{\emph{ACL}}.
\newblock


\bibitem[Lan and Jiang(2021)]%
        {lan2021modeling}
\bibfield{author}{\bibinfo{person}{Yunshi Lan} {and} \bibinfo{person}{Jing
  Jiang}.} \bibinfo{year}{2021}\natexlab{}.
\newblock \showarticletitle{Modeling transitions of focal entities for
  conversational knowledge base question answering}. In
  \bibinfo{booktitle}{\emph{ACL}}.
\newblock


\bibitem[Lewis et~al\mbox{.}(2020)]%
        {lewis2020bart}
\bibfield{author}{\bibinfo{person}{Mike Lewis}, \bibinfo{person}{Yinhan Liu},
  \bibinfo{person}{Naman Goyal}, \bibinfo{person}{Marjan Ghazvininejad},
  \bibinfo{person}{Abdelrahman Mohamed}, \bibinfo{person}{Omer Levy},
  \bibinfo{person}{Veselin Stoyanov}, {and} \bibinfo{person}{Luke
  Zettlemoyer}.} \bibinfo{year}{2020}\natexlab{}.
\newblock \showarticletitle{{BART: D}enoising Sequence-to-Sequence Pre-training
  for Natural Language Generation, Translation, and Comprehension}. In
  \bibinfo{booktitle}{\emph{ACL}}.
\newblock


\bibitem[Li et~al\mbox{.}(2022b)]%
        {li2022cooperative}
\bibfield{author}{\bibinfo{person}{Xiangsheng Li}, \bibinfo{person}{Jiaxin
  Mao}, \bibinfo{person}{Weizhi Ma}, \bibinfo{person}{Zhijing Wu},
  \bibinfo{person}{Yiqun Liu}, \bibinfo{person}{Min Zhang},
  \bibinfo{person}{Shaoping Ma}, \bibinfo{person}{Zhaowei Wang}, {and}
  \bibinfo{person}{Xiuqiang He}.} \bibinfo{year}{2022}\natexlab{b}.
\newblock \showarticletitle{A Cooperative Neural Information Retrieval Pipeline
  with Knowledge Enhanced Automatic Query Reformulation}. In
  \bibinfo{booktitle}{\emph{WSDM}}.
\newblock


\bibitem[Li et~al\mbox{.}(2022a)]%
        {li2022mmcoqa}
\bibfield{author}{\bibinfo{person}{Yongqi Li}, \bibinfo{person}{Wenjie Li},
  {and} \bibinfo{person}{Liqiang Nie}.} \bibinfo{year}{2022}\natexlab{a}.
\newblock \showarticletitle{{MMCoQA: Conversational Question Answering over
  Text, Tables, and Images}}. In \bibinfo{booktitle}{\emph{ACL}}.
\newblock


\bibitem[Lin et~al\mbox{.}(2021)]%
        {lin2021multi}
\bibfield{author}{\bibinfo{person}{Sheng-Chieh Lin},
  \bibinfo{person}{Jheng-Hong Yang}, \bibinfo{person}{Rodrigo Nogueira},
  \bibinfo{person}{Ming-Feng Tsai}, \bibinfo{person}{Chuan-Ju Wang}, {and}
  \bibinfo{person}{Jimmy Lin}.} \bibinfo{year}{2021}\natexlab{}.
\newblock \showarticletitle{Multi-stage conversational passage retrieval: An
  approach to fusing term importance estimation and neural query rewriting}.
\newblock \bibinfo{journal}{\emph{TOIS}} (\bibinfo{year}{2021}).
\newblock


\bibitem[Linjordet and Balog(2022)]%
        {linjordet2022would}
\bibfield{author}{\bibinfo{person}{Trond Linjordet} {and}
  \bibinfo{person}{Krisztian Balog}.} \bibinfo{year}{2022}\natexlab{}.
\newblock \showarticletitle{{Would You Ask it that Way? Measuring and Improving
  Question Naturalness for Knowledge Graph Question Answering}}. In
  \bibinfo{booktitle}{\emph{SIGIR}}.
\newblock


\bibitem[Liu et~al\mbox{.}(2020)]%
        {liu2020tell}
\bibfield{author}{\bibinfo{person}{Dayiheng Liu}, \bibinfo{person}{Yeyun Gong},
  \bibinfo{person}{Jie Fu}, \bibinfo{person}{Yu Yan}, \bibinfo{person}{Jiusheng
  Chen}, \bibinfo{person}{Jiancheng Lv}, \bibinfo{person}{Nan Duan}, {and}
  \bibinfo{person}{Ming Zhou}.} \bibinfo{year}{2020}\natexlab{}.
\newblock \showarticletitle{Tell Me How to Ask Again: Question Data
  Augmentation with Controllable Rewriting in Continuous Space}. In
  \bibinfo{booktitle}{\emph{EMNLP}}.
\newblock


\bibitem[Liu et~al\mbox{.}(2022)]%
        {liu2022rainier}
\bibfield{author}{\bibinfo{person}{Jiacheng Liu}, \bibinfo{person}{Skyler
  Hallinan}, \bibinfo{person}{Ximing Lu}, \bibinfo{person}{Pengfei He},
  \bibinfo{person}{Sean Welleck}, \bibinfo{person}{Hannaneh Hajishirzi}, {and}
  \bibinfo{person}{Yejin Choi}.} \bibinfo{year}{2022}\natexlab{}.
\newblock \showarticletitle{Rainier: Reinforced knowledge introspector for
  commonsense question answering}.
\newblock \bibinfo{journal}{\emph{arXiv}} (\bibinfo{year}{2022}).
\newblock


\bibitem[Liu et~al\mbox{.}(2019)]%
        {liu2019generative}
\bibfield{author}{\bibinfo{person}{Ye Liu}, \bibinfo{person}{Chenwei Zhang},
  \bibinfo{person}{Xiaohui Yan}, \bibinfo{person}{Yi Chang}, {and}
  \bibinfo{person}{Philip~S Yu}.} \bibinfo{year}{2019}\natexlab{}.
\newblock \showarticletitle{Generative question refinement with deep
  reinforcement learning in retrieval-based QA system}. In
  \bibinfo{booktitle}{\emph{CIKM}}.
\newblock


\bibitem[Liu and Belkin(2008)]%
        {liu2008query}
\bibfield{author}{\bibinfo{person}{Ying-Hsang Liu} {and}
  \bibinfo{person}{Nicholas~J Belkin}.} \bibinfo{year}{2008}\natexlab{}.
\newblock \showarticletitle{Query reformulation, search performance, and term
  suggestion devices in question-answering tasks}. In
  \bibinfo{booktitle}{\emph{IIiX}}.
\newblock


\bibitem[Marion et~al\mbox{.}(2021)]%
        {marion2021structured}
\bibfield{author}{\bibinfo{person}{Pierre Marion}, \bibinfo{person}{Pawel
  Nowak}, {and} \bibinfo{person}{Francesco Piccinno}.}
  \bibinfo{year}{2021}\natexlab{}.
\newblock \showarticletitle{Structured Context and High-Coverage Grammar for
  Conversational Question Answering over Knowledge Graphs}. In
  \bibinfo{booktitle}{\emph{EMNLP}}.
\newblock


\bibitem[Mnih et~al\mbox{.}(2015)]%
        {mnih2015human}
\bibfield{author}{\bibinfo{person}{Volodymyr Mnih}, \bibinfo{person}{Koray
  Kavukcuoglu}, \bibinfo{person}{David Silver}, \bibinfo{person}{Andrei~A
  Rusu}, \bibinfo{person}{Joel Veness}, \bibinfo{person}{Marc~G Bellemare},
  \bibinfo{person}{Alex Graves}, \bibinfo{person}{Martin Riedmiller},
  \bibinfo{person}{Andreas~K Fidjeland}, \bibinfo{person}{Georg Ostrovski},
  {et~al\mbox{.}}} \bibinfo{year}{2015}\natexlab{}.
\newblock \showarticletitle{Human-level control through deep reinforcement
  learning}.
\newblock \bibinfo{journal}{\emph{Nature}} \bibinfo{volume}{518},
  \bibinfo{number}{7540} (\bibinfo{year}{2015}).
\newblock


\bibitem[Mo et~al\mbox{.}(2023)]%
        {mo2023convgqr}
\bibfield{author}{\bibinfo{person}{Fengran Mo}, \bibinfo{person}{Kelong Mao},
  \bibinfo{person}{Yutao Zhu}, \bibinfo{person}{Yihong Wu},
  \bibinfo{person}{Kaiyu Huang}, {and} \bibinfo{person}{Jian-Yun Nie}.}
  \bibinfo{year}{2023}\natexlab{}.
\newblock \bibinfo{title}{{ConvGQR: G}enerative Query Reformulation for
  Conversational Search}.
\newblock
\newblock


\bibitem[Neeman et~al\mbox{.}(2022)]%
        {neeman2022disentqa}
\bibfield{author}{\bibinfo{person}{Ella Neeman}, \bibinfo{person}{Roee
  Aharoni}, \bibinfo{person}{Or Honovich}, \bibinfo{person}{Leshem Choshen},
  \bibinfo{person}{Idan Szpektor}, {and} \bibinfo{person}{Omri Abend}.}
  \bibinfo{year}{2022}\natexlab{}.
\newblock \showarticletitle{{DisentQA: Disentangling Parametric and Contextual
  Knowledge with Counterfactual Question Answering}}.
\newblock \bibinfo{journal}{\emph{arXiv preprint arXiv:2211.05655}}
  (\bibinfo{year}{2022}).
\newblock


\bibitem[Nogueira and Cho(2017)]%
        {nogueira2017task}
\bibfield{author}{\bibinfo{person}{Rodrigo Nogueira} {and}
  \bibinfo{person}{Kyunghyun Cho}.} \bibinfo{year}{2017}\natexlab{}.
\newblock \showarticletitle{Task-oriented query reformulation with
  reinforcement learning}. In \bibinfo{booktitle}{\emph{EMNLP}}.
\newblock


\bibitem[Owoicho et~al\mbox{.}(2023)]%
        {owoicho2023exploiting}
\bibfield{author}{\bibinfo{person}{Paul Owoicho}, \bibinfo{person}{Ivan
  Sekulic}, \bibinfo{person}{Mohammad Aliannejadi}, \bibinfo{person}{Jeffrey
  Dalton}, {and} \bibinfo{person}{Fabio Crestani}.}
  \bibinfo{year}{2023}\natexlab{}.
\newblock \showarticletitle{Exploiting Simulated User Feedback for
  Conversational Search: {R}anking, Rewriting, and Beyond}. In
  \bibinfo{booktitle}{\emph{SIGIR}}.
\newblock


\bibitem[Perez-Beltrachini et~al\mbox{.}(2023)]%
        {perez2023semantic}
\bibfield{author}{\bibinfo{person}{Laura Perez-Beltrachini},
  \bibinfo{person}{Parag Jain}, \bibinfo{person}{Emilio Monti}, {and}
  \bibinfo{person}{Mirella Lapata}.} \bibinfo{year}{2023}\natexlab{}.
\newblock \showarticletitle{Semantic Parsing for Conversational Question
  Answering over Knowledge Graphs}. In \bibinfo{booktitle}{\emph{EACL}}.
\newblock


\bibitem[Ponnusamy et~al\mbox{.}(2020)]%
        {ponnusamy2020feedback}
\bibfield{author}{\bibinfo{person}{Pragaash Ponnusamy},
  \bibinfo{person}{Alireza~Roshan Ghias}, \bibinfo{person}{Chenlei Guo}, {and}
  \bibinfo{person}{Ruhi Sarikaya}.} \bibinfo{year}{2020}\natexlab{}.
\newblock \showarticletitle{{Feedback-based self-learning in large-scale
  conversational AI agents}}. In \bibinfo{booktitle}{\emph{IAAI (AAAI
  Workshop)}}.
\newblock


\bibitem[Puri et~al\mbox{.}(2020)]%
        {puri2020training}
\bibfield{author}{\bibinfo{person}{Raul Puri}, \bibinfo{person}{Ryan Spring},
  \bibinfo{person}{Mohammad Shoeybi}, \bibinfo{person}{Mostofa Patwary}, {and}
  \bibinfo{person}{Bryan Catanzaro}.} \bibinfo{year}{2020}\natexlab{}.
\newblock \showarticletitle{Training Question Answering Models From Synthetic
  Data}. In \bibinfo{booktitle}{\emph{EMNLP}}.
\newblock


\bibitem[Qiu et~al\mbox{.}(2021)]%
        {qiu2021reinforced}
\bibfield{author}{\bibinfo{person}{Minghui Qiu}, \bibinfo{person}{Xinjing
  Huang}, \bibinfo{person}{Cen Chen}, \bibinfo{person}{Feng Ji},
  \bibinfo{person}{Chen Qu}, \bibinfo{person}{Wei Wei}, \bibinfo{person}{Jun
  Huang}, {and} \bibinfo{person}{Yin Zhang}.} \bibinfo{year}{2021}\natexlab{}.
\newblock \showarticletitle{Reinforced History Backtracking for Conversational
  Question Answering}. In \bibinfo{booktitle}{\emph{AAAI}}.
\newblock


\bibitem[Qu et~al\mbox{.}(2020)]%
        {qu2020open}
\bibfield{author}{\bibinfo{person}{Chen Qu}, \bibinfo{person}{Liu Yang},
  \bibinfo{person}{Cen Chen}, \bibinfo{person}{Minghui Qiu},
  \bibinfo{person}{W~Bruce Croft}, {and} \bibinfo{person}{Mohit Iyyer}.}
  \bibinfo{year}{2020}\natexlab{}.
\newblock \showarticletitle{Open-Retrieval Conversational Question Answering}.
  In \bibinfo{booktitle}{\emph{SIGIR}}.
\newblock


\bibitem[Qu et~al\mbox{.}(2019a)]%
        {qu2019bert}
\bibfield{author}{\bibinfo{person}{Chen Qu}, \bibinfo{person}{Liu Yang},
  \bibinfo{person}{Minghui Qiu}, \bibinfo{person}{W~Bruce Croft},
  \bibinfo{person}{Yongfeng Zhang}, {and} \bibinfo{person}{Mohit Iyyer}.}
  \bibinfo{year}{2019}\natexlab{a}.
\newblock \showarticletitle{{BERT with history answer embedding for
  conversational question answering}}. In \bibinfo{booktitle}{\emph{SIGIR}}.
\newblock


\bibitem[Qu et~al\mbox{.}(2019b)]%
        {qu2019attentive}
\bibfield{author}{\bibinfo{person}{Chen Qu}, \bibinfo{person}{Liu Yang},
  \bibinfo{person}{Minghui Qiu}, \bibinfo{person}{Yongfeng Zhang},
  \bibinfo{person}{Cen Chen}, \bibinfo{person}{W~Bruce Croft}, {and}
  \bibinfo{person}{Mohit Iyyer}.} \bibinfo{year}{2019}\natexlab{b}.
\newblock \showarticletitle{Attentive history selection for conversational
  question answering}. In \bibinfo{booktitle}{\emph{CIKM}}.
\newblock


\bibitem[Radlinski and Craswell(2017)]%
        {radlinski2017theoretical}
\bibfield{author}{\bibinfo{person}{Filip Radlinski} {and} \bibinfo{person}{Nick
  Craswell}.} \bibinfo{year}{2017}\natexlab{}.
\newblock \showarticletitle{A theoretical framework for conversational search}.
  In \bibinfo{booktitle}{\emph{CHIIR}}.
\newblock


\bibitem[Raposo et~al\mbox{.}(2022)]%
        {raposo2022question}
\bibfield{author}{\bibinfo{person}{Gon{\c{c}}alo Raposo}, \bibinfo{person}{Rui
  Ribeiro}, \bibinfo{person}{Bruno Martins}, {and} \bibinfo{person}{Lu{\'\i}sa
  Coheur}.} \bibinfo{year}{2022}\natexlab{}.
\newblock \showarticletitle{{Question rewriting? Assessing its importance for
  conversational question answering}}. In \bibinfo{booktitle}{\emph{ECIR
  2022}}.
\newblock


\bibitem[Reddy et~al\mbox{.}(2019)]%
        {reddy2019coqa}
\bibfield{author}{\bibinfo{person}{Siva Reddy}, \bibinfo{person}{Danqi Chen},
  {and} \bibinfo{person}{Christopher Manning}.}
  \bibinfo{year}{2019}\natexlab{}.
\newblock \showarticletitle{{CoQA: A} conversational question answering
  challenge}.
\newblock \bibinfo{journal}{\emph{TACL}}  \bibinfo{volume}{7}
  (\bibinfo{year}{2019}).
\newblock


\bibitem[Reimers and Gurevych(2019)]%
        {reimers2019sentence}
\bibfield{author}{\bibinfo{person}{Nils Reimers} {and} \bibinfo{person}{Iryna
  Gurevych}.} \bibinfo{year}{2019}\natexlab{}.
\newblock \showarticletitle{Sentence-BERT: Sentence Embeddings using Siamese
  BERT-Networks}. In \bibinfo{booktitle}{\emph{EMNLP-IJCNLP}}.
\newblock


\bibitem[Roy and Anand(2022)]%
        {saharoy2022question}
\bibfield{author}{\bibinfo{person}{Rishiraj~Saha Roy} {and}
  \bibinfo{person}{Avishek Anand}.} \bibinfo{year}{2022}\natexlab{}.
\newblock \showarticletitle{Question Answering for the Curated Web: Tasks and
  Methods in QA over Knowledge Bases and Text Collections}.
\newblock \bibinfo{journal}{\emph{Synthesis Lectures on Information Concepts,
  Retrieval, and Services}} (\bibinfo{year}{2022}).
\newblock


\bibitem[Sachan and Xing(2018)]%
        {sachan2018self}
\bibfield{author}{\bibinfo{person}{Mrinmaya Sachan} {and} \bibinfo{person}{Eric
  Xing}.} \bibinfo{year}{2018}\natexlab{}.
\newblock \showarticletitle{Self-training for jointly learning to ask and
  answer questions}. In \bibinfo{booktitle}{\emph{NAACL}}.
\newblock


\bibitem[Saha et~al\mbox{.}(2018)]%
        {saha2018complex}
\bibfield{author}{\bibinfo{person}{Amrita Saha}, \bibinfo{person}{Vardaan
  Pahuja}, \bibinfo{person}{Mitesh~M Khapra}, \bibinfo{person}{Karthik
  Sankaranarayanan}, {and} \bibinfo{person}{Sarath Chandar}.}
  \bibinfo{year}{2018}\natexlab{}.
\newblock \showarticletitle{{Complex sequential question answering: Towards
  learning to converse over linked question answer pairs with a knowledge
  graph}}. In \bibinfo{booktitle}{\emph{AAAI}}.
\newblock


\bibitem[Shen et~al\mbox{.}(2019)]%
        {shen2019multi}
\bibfield{author}{\bibinfo{person}{Tao Shen}, \bibinfo{person}{Xiubo Geng},
  \bibinfo{person}{Tao Qin}, \bibinfo{person}{Daya Guo}, \bibinfo{person}{Duyu
  Tang}, \bibinfo{person}{Nan Duan}, \bibinfo{person}{Guodong Long}, {and}
  \bibinfo{person}{Daxin Jiang}.} \bibinfo{year}{2019}\natexlab{}.
\newblock \showarticletitle{Multi-Task Learning for Conversational Question
  Answering over a Large-Scale Knowledge Base}. In
  \bibinfo{booktitle}{\emph{EMNLP}}.
\newblock


\bibitem[Shen et~al\mbox{.}(2022)]%
        {shen2022product}
\bibfield{author}{\bibinfo{person}{Xiaoyu Shen}, \bibinfo{person}{Gianni
  Barlacchi}, \bibinfo{person}{Marco Del~Tredici}, \bibinfo{person}{Weiwei
  Cheng}, \bibinfo{person}{Bill Byrne}, {and} \bibinfo{person}{Adri{\`a} de
  Gispert}.} \bibinfo{year}{2022}\natexlab{}.
\newblock \showarticletitle{{Product Answer Generation from Heterogeneous
  Sources: A New Benchmark and Best Practices}}. In
  \bibinfo{booktitle}{\emph{ECNLP}}.
\newblock


\bibitem[Suchanek et~al\mbox{.}(2007)]%
        {suchanek2007yago}
\bibfield{author}{\bibinfo{person}{Fabian Suchanek}, \bibinfo{person}{Gjergji
  Kasneci}, {and} \bibinfo{person}{Gerhard Weikum}.}
  \bibinfo{year}{2007}\natexlab{}.
\newblock \showarticletitle{{YAGO: A core of semantic knowledge}}. In
  \bibinfo{booktitle}{\emph{WWW}}.
\newblock


\bibitem[Sun et~al\mbox{.}(2023)]%
        {sun2023history}
\bibfield{author}{\bibinfo{person}{Hao Sun}, \bibinfo{person}{Yang Li},
  \bibinfo{person}{Liwei Deng}, \bibinfo{person}{Bowen Li},
  \bibinfo{person}{Binyuan Hui}, \bibinfo{person}{Binhua Li},
  \bibinfo{person}{Yunshi Lan}, \bibinfo{person}{Yan Zhang}, {and}
  \bibinfo{person}{Yongbin Li}.} \bibinfo{year}{2023}\natexlab{}.
\newblock \showarticletitle{History Semantic Graph Enhanced Conversational
  {KBQA} with Temporal Information Modeling}. In
  \bibinfo{booktitle}{\emph{ACL}}.
\newblock


\bibitem[Tomuro(2003)]%
        {tomuro2003interrogative}
\bibfield{author}{\bibinfo{person}{Noriko Tomuro}.}
  \bibinfo{year}{2003}\natexlab{}.
\newblock \showarticletitle{Interrogative reformulation patterns and
  acquisition of question paraphrases}. In
  \bibinfo{booktitle}{\emph{Proceedings of the second international workshop on
  Paraphrasing}}.
\newblock


\bibitem[Tomuro and Lytinen(2001)]%
        {tomuro2001selecting}
\bibfield{author}{\bibinfo{person}{Noriko Tomuro} {and}
  \bibinfo{person}{Steven~L. Lytinen}.} \bibinfo{year}{2001}\natexlab{}.
\newblock \showarticletitle{Selecting features for paraphrasing question
  sentences}. In \bibinfo{booktitle}{\emph{NLPRS}}.
\newblock


\bibitem[Usbeck et~al\mbox{.}(2017)]%
        {usbeck2017open}
\bibfield{author}{\bibinfo{person}{Ricardo Usbeck},
  \bibinfo{person}{Axel-Cyrille~Ngonga Ngomo}, \bibinfo{person}{Bastian
  Haarmann}, \bibinfo{person}{Anastasia Krithara}, \bibinfo{person}{Michael
  R{\"o}der}, {and} \bibinfo{person}{Giulio Napolitano}.}
  \bibinfo{year}{2017}\natexlab{}.
\newblock \showarticletitle{{7th Open challenge on question answering over
  linked data (QALD-7)}}. In \bibinfo{booktitle}{\emph{Semantic Web Evaluation
  Challenge}}.
\newblock


\bibitem[Vakulenko et~al\mbox{.}(2020)]%
        {vakulenko2020wrong}
\bibfield{author}{\bibinfo{person}{Svitlana Vakulenko}, \bibinfo{person}{Shayne
  Longpre}, \bibinfo{person}{Zhucheng Tu}, {and} \bibinfo{person}{Raviteja
  Anantha}.} \bibinfo{year}{2020}\natexlab{}.
\newblock \showarticletitle{A Wrong Answer or a Wrong Question? An Intricate
  Relationship between Question Reformulation and Answer Selection in
  Conversational Question Answering}. In \bibinfo{booktitle}{\emph{SCAI}}.
\newblock


\bibitem[Vakulenko et~al\mbox{.}(2021)]%
        {vakulenko2021question}
\bibfield{author}{\bibinfo{person}{Svitlana Vakulenko}, \bibinfo{person}{Shayne
  Longpre}, \bibinfo{person}{Zhucheng Tu}, {and} \bibinfo{person}{Raviteja
  Anantha}.} \bibinfo{year}{2021}\natexlab{}.
\newblock \showarticletitle{Question rewriting for conversational question
  answering}. In \bibinfo{booktitle}{\emph{WSDM}}.
\newblock


\bibitem[Voorhees(1999)]%
        {voorhees1999trec}
\bibfield{author}{\bibinfo{person}{Ellen~M. Voorhees}.}
  \bibinfo{year}{1999}\natexlab{}.
\newblock \showarticletitle{{The TREC-8 question answering track report}}. In
  \bibinfo{booktitle}{\emph{TREC}}.
\newblock


\bibitem[Voskarides et~al\mbox{.}(2020)]%
        {voskarides2020query}
\bibfield{author}{\bibinfo{person}{Nikos Voskarides}, \bibinfo{person}{Dan Li},
  \bibinfo{person}{Pengjie Ren}, \bibinfo{person}{Evangelos Kanoulas}, {and}
  \bibinfo{person}{Maarten de Rijke}.} \bibinfo{year}{2020}\natexlab{}.
\newblock \showarticletitle{Query resolution for conversational search with
  limited supervision}. In \bibinfo{booktitle}{\emph{SIGIR}}.
\newblock


\bibitem[Vrande{\v{c}}i{\'c} and Kr{\"o}tzsch(2014)]%
        {vrandevcic2014wikidata}
\bibfield{author}{\bibinfo{person}{Denny Vrande{\v{c}}i{\'c}} {and}
  \bibinfo{person}{Markus Kr{\"o}tzsch}.} \bibinfo{year}{2014}\natexlab{}.
\newblock \showarticletitle{{Wikidata: A free collaborative knowledge base}}.
\newblock \bibinfo{journal}{\emph{CACM}} \bibinfo{volume}{57},
  \bibinfo{number}{10} (\bibinfo{year}{2014}).
\newblock


\bibitem[Williams(1992)]%
        {williams1992simple}
\bibfield{author}{\bibinfo{person}{Ronald~J Williams}.}
  \bibinfo{year}{1992}\natexlab{}.
\newblock \showarticletitle{Simple statistical gradient-following algorithms
  for connectionist reinforcement learning}.
\newblock \bibinfo{journal}{\emph{Machine learning}} \bibinfo{volume}{8},
  \bibinfo{number}{3-4} (\bibinfo{year}{1992}).
\newblock


\bibitem[Xue et~al\mbox{.}(2012)]%
        {xue2012automatically}
\bibfield{author}{\bibinfo{person}{Xiaobing Xue}, \bibinfo{person}{Yu Tao},
  \bibinfo{person}{Daxin Jiang}, {and} \bibinfo{person}{Hang Li}.}
  \bibinfo{year}{2012}\natexlab{}.
\newblock \showarticletitle{Automatically mining question reformulation
  patterns from search log data}. In \bibinfo{booktitle}{\emph{ACL}}.
\newblock


\bibitem[Yang et~al\mbox{.}(2019)]%
        {yang2019data}
\bibfield{author}{\bibinfo{person}{Wei Yang}, \bibinfo{person}{Yuqing Xie},
  \bibinfo{person}{Luchen Tan}, \bibinfo{person}{Kun Xiong},
  \bibinfo{person}{Ming Li}, {and} \bibinfo{person}{Jimmy Lin}.}
  \bibinfo{year}{2019}\natexlab{}.
\newblock \showarticletitle{Data augmentation for bert fine-tuning in
  open-domain question answering}.
\newblock \bibinfo{journal}{\emph{arXiv preprint arXiv:1904.06652}}
  (\bibinfo{year}{2019}).
\newblock


\bibitem[Yih et~al\mbox{.}(2015)]%
        {yih2015semantic}
\bibfield{author}{\bibinfo{person}{W. Yih}, \bibinfo{person}{M. Chang},
  \bibinfo{person}{X. He}, {and} \bibinfo{person}{J. Gao}.}
  \bibinfo{year}{2015}\natexlab{}.
\newblock \showarticletitle{Semantic Parsing via Staged Query Graph Generation:
  Question Answering with Knowledge Base}. In \bibinfo{booktitle}{\emph{ACL}}.
\newblock


\bibitem[Yu et~al\mbox{.}(2020)]%
        {yu2020few}
\bibfield{author}{\bibinfo{person}{Shi Yu}, \bibinfo{person}{Jiahua Liu},
  \bibinfo{person}{Jingqin Yang}, \bibinfo{person}{Chenyan Xiong},
  \bibinfo{person}{Paul Bennett}, \bibinfo{person}{Jianfeng Gao}, {and}
  \bibinfo{person}{Zhiyuan Liu}.} \bibinfo{year}{2020}\natexlab{}.
\newblock \showarticletitle{Few-Shot Generative Conversational Query
  Rewriting}. In \bibinfo{booktitle}{\emph{SIGIR}}.
\newblock


\bibitem[Zamani et~al\mbox{.}(2022)]%
        {zamani2022conversational}
\bibfield{author}{\bibinfo{person}{Hamed Zamani}, \bibinfo{person}{Johanne~R
  Trippas}, \bibinfo{person}{Jeff Dalton}, {and} \bibinfo{person}{Filip
  Radlinski}.} \bibinfo{year}{2022}\natexlab{}.
\newblock \showarticletitle{Conversational information seeking}.
\newblock \bibinfo{journal}{\emph{arXiv preprint arXiv:2201.08808}}
  (\bibinfo{year}{2022}).
\newblock


\bibitem[Zhang et~al\mbox{.}(2022)]%
        {zhang2022analyzing}
\bibfield{author}{\bibinfo{person}{Shuo Zhang}, \bibinfo{person}{Mu-Chun Wang},
  {and} \bibinfo{person}{Krisztian Balog}.} \bibinfo{year}{2022}\natexlab{}.
\newblock \showarticletitle{Analyzing and Simulating User Utterance
  Reformulation in Conversational Recommender Systems}. In
  \bibinfo{booktitle}{\emph{SIGIR}}.
\newblock


\end{thebibliography}
